\documentclass[submission, Phys]{SciPost}

\usepackage[utf8]{inputenc} %
\usepackage[T1]{fontenc} 	%
\usepackage[english]{babel} %

\usepackage[bitstream-charter]{mathdesign}
\usepackage{geometry} 		%
\usepackage{amsmath} 		%
\usepackage{mathtools} 		%
\usepackage{float} 			%
\usepackage{graphicx} 		%
\usepackage{tabularx} 		%
\usepackage{booktabs} 		%
\usepackage{color, xcolor} 	%
\usepackage{pdfpages} 		%
\usepackage{extarrows} 		%
\usepackage{multirow} 		%
\usepackage{multicol} 		%
\usepackage{enumitem} 		%
\usepackage{xspace} 		%
\usepackage{stackrel} 		%
\usepackage{tikz} 			%
\usepackage{braket} 		%
\usepackage{bm} 			%
\usepackage{tensor} 		%
\usepackage{slashed} 		%
\usepackage{siunitx} 		%
\usepackage{lastpage} 		%
\usepackage{cite} 			%
\usepackage[normalem]{ulem} %
\usepackage{fontawesome} 	%
\usepackage{tocloft} 		%
\usepackage{titlesec} 		%
\usepackage{doi} 			%
\usepackage{hyperref} 		%
\usepackage[most]{tcolorbox} 					%
\usepackage[nameinlink, capitalize]{cleveref} 	%
\usepackage[nottoc, notlot, notlof]{tocbibind} 	%
\usepackage[ruled, vlined]{algorithm2e} 		%
\usepackage{makecell}
\usepackage[makeroom]{cancel}
\usepackage{feynmf}

\makeatletter
\def\BState{\State\hskip-\ALG@thistlm}
\makeatother

\makeatletter
\@ifundefined{pdfoutput}{}{\DeclareGraphicsRule{*}{mps}{*}{}}
\makeatother

\makeatletter
\DeclareRobustCommand*{\bfseries}{%
   \not@math@alphabet\bfseries\mathbf
   \fontseries\bfdefault\selectfont
   \boldmath
}
\makeatother

\hypersetup{
	pdftitle={},
	pdfauthor={},
	colorlinks=true, 			%
	linkcolor={red!50!black}, 	%
	citecolor={blue!50!black}, 	%
	urlcolor={blue!80!black} 	%
} 

\DeclareSymbolFont{usualmathcal}{OMS}{cmsy}{m}{n}
\DeclareSymbolFontAlphabet{\mathcal}{usualmathcal}

\fancypagestyle{SPstyle}{
\fancyhf{}
\lhead{\colorbox{scipostblue}{\bf \color{white} ~SciPost Physics Community Reports }}
\rhead{{\bf \color{scipostdeepblue} ~Submission }}

\fancyfoot[C]{\textbf{\thepage}}
}

\SetArgSty{textnormal}
\SetKwComment{Comment}{{\small\#}~}{}
\SetCommentSty{mycommfont}

\setitemize{itemsep=0pt, parsep=0pt} 				%
\setenumerate{itemsep=0pt, parsep=0pt} 				%
\setitemize{itemsep=2pt,topsep=2pt,parsep=0pt,partopsep=0pt,leftmargin=*}
\setenumerate{itemsep=0pt,topsep=2pt,parsep=0pt,partopsep=0pt,labelindent=3pt,leftmargin=*}
\newlist{todolist}{itemize}{2}
\setlist[todolist]{label=$\square$}
\usepackage{pifont}

\usepackage{amsmath}

\usepackage{amsthm} 		%
\theoremstyle{definition}

\definecolor{red_cb}{HTML}{e41a1c}
\definecolor{blue_cb}{HTML}{377eb8}
\definecolor{green_cb}{HTML}{4daf4a}
\definecolor{purple_cb}{HTML}{984ea3}
\definecolor{orange_cb}{HTML}{ff7f00}

\definecolor{EmeraldGreen}{HTML}{1ea78d}
\definecolor{EnglishRed}{HTML}{b02427}
\hypersetup{colorlinks=true,urlcolor=EmeraldGreen,citecolor=EmeraldGreen,linkcolor=EnglishRed}

\newcommand{\mwith}{\text{with}}
\newcommand{\mand}{\text{and}}
\newcommand{\mfor}{\text{for}}

\def\d{\mathrm{d}}

\newcommand\one{\leavevmode\hbox{\small1\normalsize\kern-.33em1}}

\newcommand{\arXiv}[2][]{%
	\ifthenelse{\equal{#1}{}}%
	{\href{http://arxiv.org/abs/#2}{arXiv:#2}}%
	{\href{http://arxiv.org/abs/#2}{arXiv:#2~[#1]}}}

\def\slashchar#1{\setbox0=\hbox{$#1$}           %
   \dimen0=\wd0                                 %
   \setbox1=\hbox{/} \dimen1=\wd1               %
   \ifdim\dimen0>\dimen1                        %
      \rlap{\hbox to \dimen0{\hfil/\hfil}}      %
      #1                                        %
   \else                                        %
      \rlap{\hbox to \dimen1{\hfil$#1$\hfil}}   %
      /                                         %
   \fi}

\newcommand{\tikznode}[2]{%
\ifmmode%
\tikz[remember picture,baseline=(#1.base),inner sep=0pt] \node (#1) {$#2$};%
\else
\tikz[remember picture,baseline=(#1.base),inner sep=0pt] \node (#1) {#2};%
\fi}

\DeclareMathOperator*{\argmax}{arg\,max}

\graphicspath{{./figs/}}

\begin{document}

\pagestyle{SPstyle}

\begin{center}{\Large \textbf{\color{scipostdeepblue}{
An Introduction to Bayesian and Frequentist Simulation-Based Inference with Machine Learning
}}}

Part of the VERaiPHY Initiative
\end{center}

\begin{center}\textbf{
Maximilian Dax\textsuperscript{$\star$,1},
Theo Heimel\textsuperscript{$\star$,2},
and Gilles Louppe\textsuperscript{$\circ$,3}
}\end{center}

\begin{center}
{\bf 1} ELLIS Institute Tübingen, Max Planck Institute for Intelligent Systems, Tübingen AI Center, Tübingen, Germany
\\
{\bf 2} CP3, Universit\'e catholique de Louvain, Louvain-la-Neuve, Belgium
\\
{\bf 3} Montefiore Institute, University of Liège, Belgium

$\star$ {\small Leading authors}\quad
$\circ$ {\small Advisor}
\end{center}

\vspace{-2.5em}

\section*{\color{scipostdeepblue}{Abstract}}
\textbf{\boldmath{%
Simulation-based inference (SBI) with machine learning is an increasingly important tool for solving inverse problems in science and engineering, including parameter inference and the inversion of detector effects. We provide an overview of the Bayesian and frequentist statistical frameworks, describe how machine-learning-based SBI methods, such as neural posterior estimation and neural likelihood estimation, can be used for parameter estimation within these frameworks, and show that the same methods can also be applied to Empirical Bayes or unfolding tasks. We also discuss how to validate inference results and the limitations of SBI with machine learning.
}}

\vspace{6pt}
\noindent\rule{\textwidth}{1pt}
\tableofcontents

\clearpage
\section{Introduction}
\label{sec:intro}

Inverse problems arise across science and engineering, such as inferring model parameters from data or inverting detector effects. In most of these problems, there is a known forward map, expressed via a likelihood. In this review, we consider likelihoods directly mapping the model parameters to the observations,
\begin{equation}
    \label{eq:forward_model}
    \text{parameters }\mu \quad\xrightarrow[p(x|\mu)]{\text{likelihood}}\quad \text{observations }x \;,
\end{equation}
as well as scenarios where the likelihood factorizes into multiple steps,
\begin{equation}
    \label{eq:forward_model_latent}
    \text{parameters }\mu \quad\xrightarrow[p(z|\mu)]{\text{likelihood}}\quad \text{latent space }z
    \quad\xrightarrow[p(x|z)]{\text{likelihood}}\quad \text{observations }x \;.
\end{equation}
For instance, the mapping from parameters $\mu$ to the latent space $z$ might be a first-principles description of the underlying physics, whereas the mapping from $z$ to $x$ captures effects of the experimental setup.

In practice, the forward map may be implemented explicitly as a density (possibly known only up to a multiplicative normalization constant), or implicitly via a simulator that enables sampling $x\sim p(x|\mu)$.
This gives rise to two inference paradigms.
\textbf{Likelihood-based} inference is applicable whenever the likelihood can be evaluated. Common methods include Markov chain Monte Carlo (MCMC)~\cite{metropolis1953equation,Hastings:1970} and nested sampling~\cite{Skilling:2006}.
\textbf{Simulation-based} (or likelihood-free) inference~\cite{Cranmer:2019eaq} is applicable whenever we can generate data space (or latent space) samples given the model parameters.

The forward model is often not uniquely invertible, for example, due to parameter degeneracies or measurement noise. Consequently, inference must be formulated probabilistically to account for uncertainties. In this probabilistic description, we can either treat the model parameters as random variables and consider the observation fixed, or treat the observation as random and the true parameters as fixed. These two statistical frameworks are known as \textbf{Bayesian} and \textbf{frequentist} inference, respectively. In Bayesian statistics, probability distributions over the parameters $\mu$ quantify the degree of belief, and the observed data is used to update this belief. In contrast, frequentist statistics aims to provide coverage guarantees of the predicted results across multiple (hypothetical) repetitions of the experiment with independent $x$.

In this article, we describe computational techniques to perform and validate scientific inference, focusing on the simulation-based paradigm.
We first review the underlying statistical frameworks, contrasting the Bayesian and frequentist perspectives (Sec.~\ref{sec:stat-framework}), then describe machine learning techniques for simulation-based inference (Sec.~\ref{sec:ml_model}) and finally discuss methods to validate and verify inference results (Sec.~\ref{sec:verification}).

We provide a technical introduction to methodology rather than a survey of applications. For overviews of the rapidly growing literature on simulation-based inference applications, we refer the reader to Ref.~\cite{Cranmer:2019eaq} and to domain-specific reviews in particle physics~\cite{Brehmer:2020cvb} and in astrophysics and cosmology~\cite{Thiele:2026hxu}, as well as to community-maintained collections of SBI literature~\cite{sbi_website,awesome_neural_sbi}. Ref.~\cite{deistler2025simulation} further provides practical guidance and worked examples for applying simulation-based inference.

\textit{This article contributes to VERaiPHY (Validation \& Evaluation for Robust AI in PHYsics), a PHYSTAT review series establishing verification and validation standards for machine learning across particle physics, astrophysics, and cosmology.}

\section{Statistical framework}\label{sec:stat-framework}

In this section, we first outline parameter inference in the Bayesian (Sec.~\ref{subsec:stat-bayes}) and frequentist (Sec.~\ref{subsec:stat-freq}) frameworks. We then discuss how inverse problems can also be defined at the level of distributions rather than individual observations, leading to the classes of Empirical Bayes and unfolding methods (Sec.~\ref{subsec:stat-emp-bayes}). Lastly, we show how nuisance parameters can be used to account for mismodeling in the forward mapping (Sec.~\ref{subsec:stat-nuisance}).

\subsection{Bayesian inference}\label{subsec:stat-bayes}
\begin{figure}
    \centering
    \includegraphics[width=0.49\linewidth]{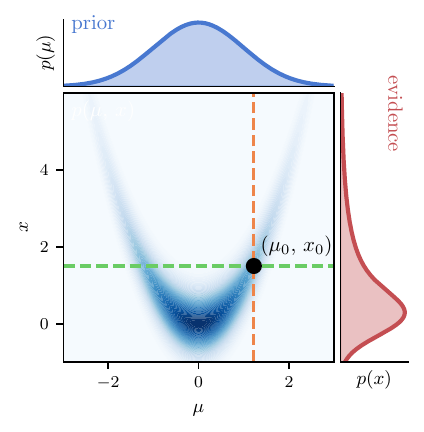}
    \includegraphics[width=0.49\linewidth]{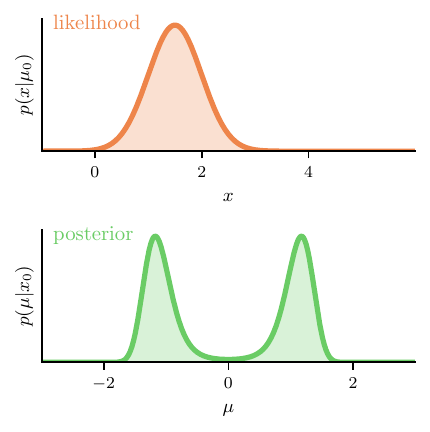}
    \caption{
    The likelihood $p(x|\mu)$ (orange) and prior $p(\mu)$ (blue) together define the joint distribution $p(\mu,x)$. Bayesian inference aims to sample the posterior $\mu\sim p(\mu|x_0)$ (green; horizontal slice through joint distribution) or evaluate the evidence $p(x_0)$ (red; horizontal projection of joint distribution) for an observation $x_0$.
    }
    \label{fig:Bayes}
\end{figure}

Bayesian inference combines prior knowledge about the parameters, encoded in the prior $p(\mu)$, with information provided by observed data, encoded in the likelihood $p(x|\mu)$.
The Bayesian framework is well suited when there is meaningful prior information about the parameters, for instance from theoretical considerations or other experiments. Such physically motivated priors are particularly useful to regularize the posterior when inference is based on only one or a small number of observations. In such cases, there is also no ensemble of identical experiments to define coverage over, making frequentist inference a less suitable description compared to the Bayesian framework where there is no notion of repeated experiments. Bayesian inference is particularly popular in astrophysics, where events are often observed only once, but substantial prior knowledge is available. For example, when characterizing an astrophysical event based on a measurement, one can encode the assumption of an isotropic distribution of sources in volume through a prior $p(\mu)\propto d^2$ quadratic in the distance $d$ from Earth. One could also integrate detailed knowledge about the spatial distribution of galaxies (see Fig.~2 in Ref.~\cite{coulter2017swope} for an example of how prior knowledge of galaxy locations combined with a skymap inferred from measured gravitational-wave data enabled detection of the famous kilonova AT 2017gfo). 

This is achieved through the \textbf{posterior} distribution, which is defined via Bayes’ theorem,
\begin{equation}\label{eq:bayes}
    p(\mu|x)=\frac{p(\mu)\,p(x|\mu)}{p(x)}.
\end{equation}
Intuitively, the prior quantifies our initial belief about $\mu$, the likelihood assesses how well $x$ fits $\mu$, and the posterior then summarizes the updated knowledge about $\mu$ after measuring $x$. In the Bayesian picture, the parameters are treated as random variables described by probability distributions, whereas the observations are treated as fixed. All these distributions can be interpreted as either slices through or projections of the joint distribution $p(\mu,x)$ (Fig.~\ref{fig:Bayes}). In addition to the posterior distribution, the result of Bayesian inference can be given in the form of expectation values of parameters and \textbf{credible intervals} or (in higher dimensions) \textbf{credible sets} $C_\text{cred}(x)$ that the parameter will fall in with a given probability $\alpha$ conditional on a fixed observation $x_0$,
\begin{equation}
    P_\mu\left\{\mu \in C_\text{cred}(x_0)\right\} = \alpha \;.
\end{equation}
The central task in Bayesian inference is then to represent the posterior through a set of samples $(\mu_1,\ldots,\mu_n)\sim p(\mu|x)$. This is a particularly useful representation, as it enables straightforward marginalization as well as estimation of expectations and credible sets. However, sampling is generally a challenging task. In likelihood-based inference, one usually employs stochastic methods such as MCMC, which are based on evaluations of the unnormalized posterior density $p(x|\mu)p(\mu)$. In the simulation-based context, one commonly trains deep neural networks to directly estimate the posterior or related distributions based on samples from the joint distribution (Sec.~\ref{sec:ml_model}).

In addition to sampling the posterior, one sometimes also needs to estimate the normalization constant in Eq.~\eqref{eq:bayes}, known as the \textbf{Bayesian evidence}. It is defined as
\begin{equation}\label{eq:evidence}
    p(x)=\int p(\mu)\,p(x|\mu)\,\d\mu.
\end{equation}
The evidence measures the plausibility of the observed data under the assumed prior and likelihood.
While it is difficult to interpret its absolute value, evidence ratios (also called Bayes factors) can be used for model comparison and selection. For example, given two forward models with associated likelihoods $p_a(x|\mu)$ and $p_b(x|\mu)$, the ratio $p_a(x)/p_b(x)$ quantifies how much more likely it is to obtain measurement $x$ with model $a$ than with $b$.
While the integral in Eq.~\eqref{eq:evidence} is generally hard to compute, some inference methods can provide unbiased estimates of the Bayesian evidence in addition to posterior samples~\cite{Skilling:2006,Dax:2022pxd} (see also Sec.~\ref{subsec:npe-is}). Importantly, all results from Bayesian inference, including Bayes factors, are only valid for the specific choice of prior. Therefore, if no defensible prior exists or the goal is to make prior-independent statements, like for discovery claims in high-energy physics, Bayesian inference is not suitable and frequentist methods are preferred.

\subsection{Frequentist inference}\label{subsec:stat-freq}
\begin{figure}
    \centering
    \includegraphics[width=0.49\linewidth]{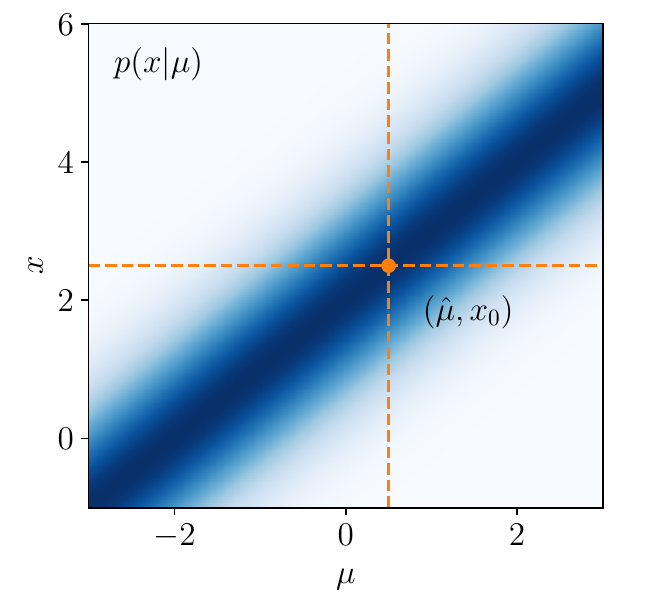}
    \includegraphics[width=0.49\linewidth]{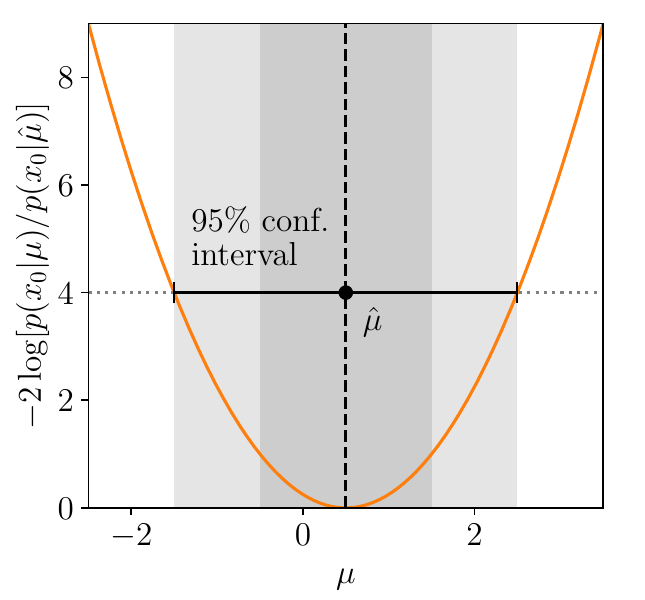}
    \caption{
    The left panel shows an example likelihood $p(x|\mu)$ with the maximum likelihood estimate $\hat\mu$ for an observation $x_0$. The right panel shows the logarithm of the likelihood ratio relative to the maximum likelihood estimate, and how a $95\%$ confidence interval can be extracted from the ratio.
    }
    \label{fig:frequentist}
\end{figure}

In contrast to the Bayesian approach, where the parameters to be inferred are treated as random variables, frequentist inference assumes the parameters are fixed and treats the observed data as random variables that vary across multiple repetitions of the experiment. The inference task is then to find \textbf{confidence intervals}, or, in higher dimensions, \textbf{confidence sets} as a function of the observed data $x$. For a given confidence level $\alpha$, confidence sets $C_\text{conf}(x)$ are defined as sets where the fixed, true value of the parameters $\mu_\text{true}$ is contained within the set with a probability of at least $\alpha$,
\begin{equation}
    P_x\left\{\mu_\text{true} \in C_\text{conf}(x)\right\} \geq \alpha \;.
\end{equation}
The set is called valid if this condition is fulfilled. Note that confidence sets are different from the previously introduced credible sets in Bayesian inference, as they assume the true parameters to be fixed, whereas the set randomly varies over multiple repetitions of the experiment. In the Bayesian case, the credible set is a function of the posterior and is fixed once it is conditioned on the observation $x_0$. In the frequentist case, the guarantee is a pre-experimental property of the procedure, not a post-hoc statement about any single interval. This is why frequentist intervals are preferred when the community needs a procedure-level guarantee, such as not claiming a discovery unless the false positive rate is below 5 sigma in high-energy physics. A common way to define confidence sets is to require the parameters to be within a contour of equal likelihood around the maximum likelihood estimate,
\begin{equation}
    \label{eq:conf_set}
    C_\text{conf}(x) = \left\{ \mu \:\big|\: p(x | \mu) \geq \exp\left(-\frac{c}{2}\right) \, p(x | \hat\mu) \right\} \qquad\mwith\qquad \hat\mu = \argmax_\mu p(x | \mu) \;,
\end{equation}
where $c$ has to be chosen such that $C_\text{conf}(x)$ has the desired confidence level. For multiple independent and identically distributed observations, the combined likelihood reads
\begin{equation}
    \label{eq:combine_likelihoods}
    p(x_1, \ldots, x_n|\mu) = p(x_1|\mu) \cdots p(x_n|\mu) \;.
\end{equation}
Extracting confidence sets becomes very simple in the limit of many independent observations according to Wilks' theorem. This theorem states that the negative log-likelihood ratio,
\begin{equation}
    -2 \log r(x|\mu, \mu_0) = -2 \log \frac{p(x|\mu)}{p(x|\mu_0)} \;,
\end{equation}
becomes $\chi^2$-distributed for $x \sim p(x|\mu_0)$. The log-likelihood as a function of $\mu$ then is a parabola. In one dimension, setting the values $c=1,4,9$ in Eq.\eqref{eq:conf_set} yields $68.3\%$, $95.5\%$ and $99.7\%$ confidence intervals, corresponding to the usual $1\sigma, 2\sigma, 3\sigma$ levels. Confidence sets are then given by
\begin{equation}
    C_\text{conf}(x) = \big\{ \mu \:\big|\: -2 \log r(x | \mu, \hat\mu) \leq c\big\} \;.
\end{equation}
We illustrate the process of extracting confidence intervals from the log-likelihood ratio in the asymptotic limit in Fig.~\ref{fig:frequentist}. These properties make the likelihood ratio and its logarithm the central tools in frequentist inference. However, outside of the asymptotic limit, the process becomes more complex. While extracting confidence intervals remains possible, for example via the Neyman construction~\cite{neyman1937outline}, it requires more computational effort.

Because of the multiplicative properties of likelihoods and likelihood ratios, and the computationally cheap inference in the asymptotic limit, frequentist inference is particularly suited for applications where large numbers of independent observations have to be combined. That makes it the natural inference framework in fields like particle physics, where the large number of observed collider events can be seen as many repetitions of the same experiment, yielding estimates of the same model parameters. We discuss the practical aspects of combining multiple observations in simulation-based inference in Sec.~\ref{subsec:combine_obs}. Moreover, frequentist inference does not involve prior distributions and therefore remains applicable even when reliable priors are unavailable.

\subsection{Distribution-level inference}\label{subsec:stat-emp-bayes}

In some cases, we are not interested in inferring the possible parameters $\mu$ for a given observation $x$, and instead want to know which distribution of parameters $p(\mu)$ or latent distribution $p(z)$ gives rise to the observed data distribution $p(x)$ for a given forward process,
\begin{equation}
    \label{eq:dist_level_inf}
    p(x) = \int p(\mu) \,p(x|\mu)\,\d\mu
    \qquad\text{or}\qquad
    p(x) = \int p(z)\, p(x|z)\,\d z \;.
\end{equation}
In contrast to Bayesian inference, $p(\mu)$ does not quantify a degree of belief. Instead, it is a statement about a physical population distribution.
Similar tasks appear in various areas of math and science under various names:
\begin{itemize}
\item \textbf{Unfolding detector effects}: In particle physics, collider events are simulated using a complex chain of tools. One of the most costly parts of this chain is to simulate detector effects. To simplify parameter inference and reduce computational cost, the data is often analyzed in a two-step procedure: First, only the detector effects are inverted, yielding a distribution of events at some latent stage of the simulation chain, $p(z)$. This latent information can then be used multiple times to measure parameters and test hypotheses without having to rerun the costly detector simulation.
\item \textbf{Empirical Bayes}: In statistics, there are some situations where very little information about the prior for some parameters is known. In these cases, it can be useful to extract the prior $p(\mu)$ from the observed data distribution $p(x)$ rather than making an uninformed decision about the prior, which can bias the results.
\item \textbf{Quantification learning} in machine learning is the task of inferring the relative frequencies of different labels in an unlabeled dataset, $p(x)$. In this case, $p(\mu)$ typically describes the probabilities of a discrete set of classes instead of a continuous probability density~\cite{gonzalez2017review}.
\item \textbf{Image deconvolution} is the task of removing noise from an image, given a forward model for the noise. The distributions $p(x)$ and $p(z)$ are then interpreted as the pixel intensity as a function of coordinates and color channels, rather than as probability distributions.
\end{itemize}
In all of these cases, the underlying task is mathematically equivalent, and similar methods can be applied. For instance, the iterative Bayesian unfolding~\cite{DAgostini:1994fjx} method in particle physics that solves the unfolding problem in low-dimensional, binned cases is equivalent to the Richardson–Lucy deconvolution for denoising images~\cite{richardson1972bayesian,lucy1974iterative}. For a parameterization of the target distribution $q_\theta(\mu)$ with trainable parameters $\theta$ and data points $x_1,\ldots,x_n$, the task is often defined as maximizing the likelihood
\begin{equation}
    \label{eq:dist_level_opt}
    \hat\theta = \argmax_{\theta} \int q_\theta(\mu) \, p(x_1,\ldots,x_n|\mu) \, \d\mu \;.
\end{equation}
As the target is to infer the distribution $p(\mu)$ instead of the parameters $\mu$ themselves, we will, in the following, refer to these methods as \textbf{distribution-level inference}.

\subsection{Nuisance parameters}\label{subsec:stat-nuisance}

\begin{figure}
    \centering
    \includegraphics[width=0.49\linewidth]{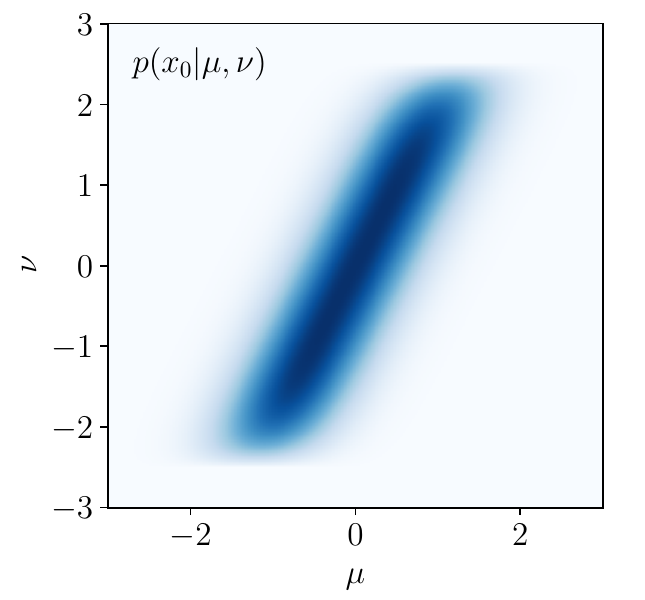}
    \includegraphics[width=0.49\linewidth]{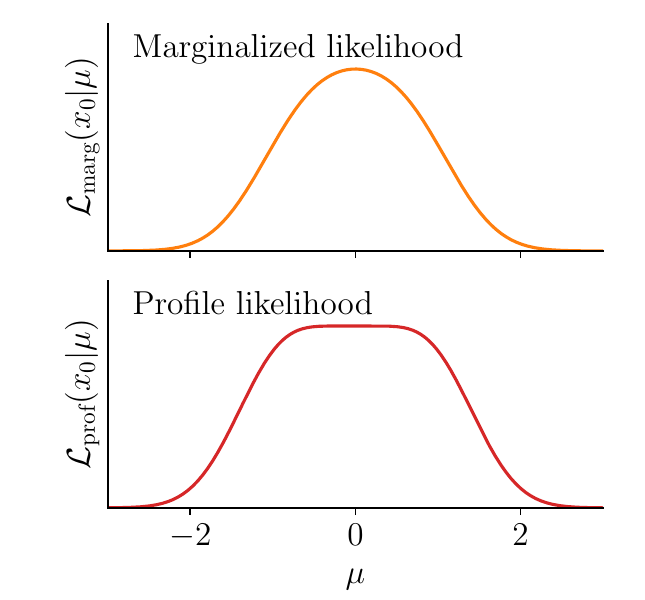}
    \caption{
    The left panel shows an example likelihood conditional on parameters of interest $\mu$ and nuisance parameters $\nu$ for an observation $x_0$. The right panels show the marginalized and profile likelihoods obtained by integrating or maximizing over the nuisance parameters.
    }
    \label{fig:nuisance}
\end{figure}

Physical models often have parameters we are not interested in measuring, yet they still influence simulation results. Such parameters are referred to as \textbf{nuisance parameters}. They allow us to absorb mismodeling of the physical truth, thereby reducing the bias of parameter estimates. Nuisance parameters arise both on the theory side, for instance, when some parts of our simulations are described by phenomenological models with parameters that have to be fitted to the data, and on the experimental side, where parameters of the experimental setup may be unknown. To perform inference, we have to absorb the effects nuisance parameters into the inference result for the parameters of interest. This is typically done using either \textbf{marginalization}, where the effect of the nuisance parameters is averaged out, or \textbf{profiling}, where the nuisance parameter is chosen for each parameter point to best describe the observed data.

Consider a likelihood $p(x|\mu, \nu)$ depending on the parameters of interest $\mu$ and nuisance parameters $\nu$. To marginalize over the nuisance parameters, we first assume a prior $p(\nu|\mu)$ that, in general, depends on the model parameters $\mu$. The \textbf{marginalized likelihood} is then defined as
\begin{equation}
    \mathcal{L}_\text{marg}(x|\mu) = \int p(\nu|\mu)\, p(x|\mu, \nu)\, \d\nu \;.
\end{equation}
Posteriors derived from the marginalized likelihood are equal to posteriors derived from the full likelihood with the nuisance parameters integrated out,
\begin{equation}
    \frac{\mathcal{L}_\text{marg}(x|\mu) \, p(\mu)}{p(x)}
    = \int \frac{p(\mu) \, p(\nu|\mu) \, p(x|\mu, \nu)}{p(x)} \,\d\nu
    = \int p(\mu, \nu|x)\,\d\nu
    = p(\mu | x) \;.
\end{equation}
This makes marginalization the natural treatment of nuisance parameters in the framework of Bayesian inference. As long as we account for nuisance parameters and their prior distributions when running our simulator, we do not need to explicitly parameterize them when sampling from the posterior, making Bayesian simulation-based inference in the presence of nuisance parameters relatively straightforward. It is, however, not always possible to perform marginalization at the simulation stage. For instance, importance sampling, as well as the computation of posterior predictive distributions and other closure tests, often requires sampling the posterior for the full set of parameters.

On the other hand, the \textbf{profile likelihood} is defined by maximizing the likelihood for given model parameters over the nuisance parameters,
\begin{equation}
    \mathcal{L}_\text{prof}(x|\mu) = \max_\nu p(x|\mu, \nu) \; .
\end{equation}
Because the profile likelihood is computed using the nuisance parameters that best describe the observed data at each parameter point $\mu$, profiling never disfavors certain $\mu$ for an unsuitable choice of nuisance parameters. This results in the most conservative possible treatment of $\mu$, in contrast to fixing $\nu$
at some other value such as its global best fit. This is in contrast to the marginalized likelihood, which treats nuisance parameters probabilistically. It is particularly suited in a frequentist setup as it does not require imposing a prior, and the maximum likelihood estimates from the profile likelihood and full likelihood coincide by definition,
\begin{equation}
    \hat\mu = \argmax_\mu \,\max_\nu p(x|\mu, \nu) = \argmax_\mu \mathcal{L}_\text{prof}(x | \mu) \;.
\end{equation}
Fig.~\ref{fig:nuisance} shows the marginalized and profile likelihood for an example distribution, where the maximization leads to a flatter shape of the profile likelihood around its maximum compared to the marginalized version. If a parameter $\mu$ is in a confidence set from the profile likelihood, $\mu \in C_\text{prof}(x)$, there is always a $\nu$ so that $(\mu,\nu)$ is in the confidence set for the full likelihood, $(\mu,\nu) \in C(x)$. This ensures the validity of the confidence intervals. It is possible to mix approaches, and for example, marginalize over nuisance parameters and then use frequentist inference for the parameters of interest. This can be advantageous in situations where credible priors are available for the nuisance parameters, or where the maximizing procedure necessary to perform profiling would be computationally prohibitive.

Nuisance parameters, however, are not always a feasible option for accounting for mismodeling in the simulation. It might not be possible to describe such mismodeling adequately using nuisance parameters, or doing so might require a large number of them. This can lead to more complex data and exploding dataset sizes. Moreover, machine-learning models for simulation-based inference become more complex and costly to train and evaluate when surrogates for distributions or likelihood ratios must be learned as functions of nuisance parameters, in addition to observations and parameters of interest. For applications with many weakly correlated nuisance parameters, as is common in particle physics, factorized approaches have been explored to address this issue by modeling the effect of individual parameters or groups of parameters with separate networks~\cite{Schofbeck:2024zjo,Barrue:2026qul,Valsecchi:2026kjy}. This results in a more favorable scaling of model and dataset size as well as training complexity.

\section{Machine learning for simulation-based inference}
\label{sec:ml_model}

Solving inverse problems is often a trade-off between computational cost and constraining power. When the likelihood is only available via simulation, classical likelihood-based methods for Bayesian and frequentist inference commonly become infeasible. The classical Bayesian approach in this setting is Approximate Bayesian Computation~\cite{pritchard1999population,beaumont2002approximate,sisson2018handbook}. Posterior samples are obtained by accepting parameters $\mu_i\sim p(\mu)$ whose simulations $x_i\sim p(x|\mu_i)$ lie within a tolerance $\varepsilon$ of the observation, $\|x_i - x_0\| < \varepsilon$. Accurate inference requires very small $\varepsilon$, in which case the acceptance rate typically becomes prohibitively low. In the frequentist setting, the likelihood itself can be estimated from simulations, for instance by histogramming simulated data. These approaches usually only remain feasible when the data is first compressed to low-dimensional (typically hand-crafted) summary statistics. Such summary statistics often lead to a loss of information, and, therefore, less strict bounds on the parameters of interest. Many modern simulation-based methods leverage deep learning to lift these restrictions and enable tractable inference.

Deep inference typically involves two stages. First, a deep neural network is trained with simulated data. This encodes information about the simulator in the neural network weights.
Second, inference for an observation of interest is performed with the trained neural network. This bypasses the need for simulations at inference time.

By decoupling training from inference, this approach front-loads the computational cost. The training stage, including simulating data and optimizing the network, is performed offline ahead of the actual inference step. Inference then merely requires cheap neural network computation. 
This enables \textbf{efficient amortized inference}: if a trained model is used for multiple observations, the cost of training is \emph{amortized} across all individual analyses.
It further enables \textbf{fast inference}, which can be critical in real-time applications where the inference result informs subsequent experimental decisions. 

While simulation-based inference has originally been developed to enable tractable inference when classical methods fail, it is nowadays also commonly applied even in contexts where conventional (likelihood-based) inference would be applicable to leverage these advantages of front-loaded, amortized computation.

In the remainder of this section, we describe how deep generative models (Sec.~\ref{subsec:ml-gen-models}) and classifiers (Sec.~\ref{subsec:ml-classifiers}) can be leveraged for simulation-based inference. We then discuss how analytically known likelihoods can be incorporated into simulation-based inference (Sec.~\ref{subsec:ml-likelihood}). Next, we extend our discussion to distribution-level inference for empirical Bayes or unfolding tasks (Sec.~\ref{subsec:ml-emp-bayes}). Finally, we show how these methods apply to different data modalities and how multiple independent observations can be combined (Sec.~\ref{subsec:ml-data-modality}).

\subsection{Density estimation}\label{subsec:ml-gen-models}
Parameter estimation is based on conditional probability densities, specifically the posterior $p(\mu|x)$ and the likelihood $p(x|\mu)$. In simulation-based inference, these are accessible through simulated samples $(\mu,x)$ from the joint distribution. Such samples can be used to train a deep generative model
\begin{equation}
    q_\theta(\mu|x)\approx p(\mu|x)
\end{equation}
for \textbf{neural posterior estimation (NPE)}~\cite{papamakarios2016fast,lueckmann2017flexible,greenberg2019automatic} or 
\begin{equation}
    q_\theta(x|\mu)\approx p(x|\mu)
\end{equation}
for \textbf{neural likelihood estimation (NLE)}~\cite{wood2010statistical,drovandi2018approximating,papamakarios2019sequential,lueckmann2019likelihood}. Once trained, these models can be used as surrogates for the respective distributions. NPE inference is particularly efficient, as it can directly sample from the trained posterior surrogate, $\mu\sim q_\theta(\mu|x)$. NLE on the other hand requires an additional sampling procedure such as MCMC to sample its posterior estimate $q_\theta(x|\mu)p(\mu)$ for Bayesian inference, making the inference not fully amortized, or a method for extracting confidence sets for frequentist inference.

The surrogate can be any model that allows sample-based training and efficient sampling (NPE) or density evaluation (NLE). This matches the standard paradigm of deep generative modeling. Indeed, many conditional generative models, including normalizing flows~\cite{rezende2015variational,papamakarios2021normalizing}, generative adversarial networks~\cite{goodfellow2014generative}, diffusion models and score matching~\cite{sohl2015deep,song2019generative,ho2020denoising}, flow matching~\cite{lipman2022flow,liuflow,albergobuilding}, and consistency models~\cite{song2023consistency}, have been adapted for simulation-based inference~\cite{greenberg2019automatic,radev2020bayesflow,ramesh2022gatsbi,geffner2022score,sharrock2022sequential,wildberger2023flow,schmitt2023consistency}. 

Normalizing flows are particularly popular for simulation-based inference. They often implement efficient density evaluation by implementing an invertible transformation with a tractable Jacobian, for example, by implementing invertible transformations using coupling blocks, where one half of the input vector is transformed as a function of the other half~\cite{dinh2016density}, or using an autoregressive structure~\cite{papamakarios2017masked}.
This allows for stable optimization via the \textbf{Kullback-Leibler divergence (KLD)} as well as verification and evidence estimation via importance sampling (Sec.~\ref{subsec:npe-is}).
The expressivity of normalizing flows is, however, limited by their restrictive architecture, leading to poor scaling to high-dimensional data. Because scientific inference is often associated with low-dimensional parameter spaces, they remain a suitable choice for many simulation-based inference tasks.

In contrast, many other generative approaches offer greater flexibility in their underlying neural network architecture, resulting in more expressive density estimation and better scaling to higher dimensions. At the same time, sampling is often computationally more costly than with normalizing flows, and evaluating the density further increases the cost or is infeasible. For instance, sampling from models trained using score matching or flow matching requires numerically integrating a differential equation, involving many calls to the neural network encoding the score or velocity field, and its gradient if the density is required~\cite{chen2018neural}.

Below, we explore the fundamentals of density estimation for simulation-based inference. In this work, we focus on normalizing flows trained with a KLD loss to provide an intuitive description of the principles underlying simulation-based inference using generative models. However, in practice, the best choice of generative architecture depends on the specific inference task and its requirements for training and inference speed, model expressivity, and density evaluation capability. We first discuss optimization of the forward KL divergence for unconditional (Sec.~\ref{subsec:KLD-uncond}) and conditional distributions (Sec.~\ref{subsec:KLD-cond}) and then discuss the training dynamics (Sec.~\ref{subsec:KLD-practical}). Finally, we briefly comment on sequential inference as a technique to tune inference to a specific observation (Sec.~\ref{subsec:sequential-inference}).

\subsubsection{The KLD for unconditional distributions}\label{subsec:KLD-uncond}

\begin{figure}
    \centering
    \includegraphics[width=0.6\linewidth]{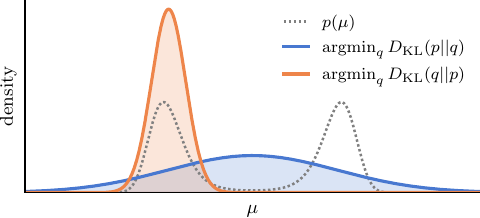}
    \caption{A Gaussian fitted to a bimodal distribution (gray) covers both modes with the forward KLD (blue) objective and only one mode with the reverse KLD (orange).
    }
    \label{fig:KLD}
\end{figure}

The KLD measures the deviation between two distributions $p(\mu)$ and $q_\theta(\mu)$. It is defined as
\begin{equation}\label{eq:KLD-definition}
    D_\mathrm{KL}(p||q_\theta) 
    = \int p(\mu)\log\left(\frac{p(\mu)}{q_\theta(\mu)}\right)\mathrm{d}\mu 
    = \mathbb{E}_{\mu\sim p(\mu)} \log\left(\frac{p(\mu)}{q_\theta(\mu)}\right).
\end{equation}
The KLD is not symmetric in its arguments. When fitting a model $q_\theta$ (e.g., a normalizing flow) to a target distribution $p$ via KLD minimization, the order determines the training dynamics. 

The \textbf{forward KLD} $D_\mathrm{KL}(p||q_\theta)$ acts as a \emph{mean-seeking} (or \emph{mass-covering}) objective. This KLD only remains finite if $q_\theta(\mu)>0$ wherever $p(\mu)>0$, so its optimization forces $q_\theta$ to place probability mass across the entire support of $p$. Heuristically, this ensures that plausible samples from $p$ are also plausible under $q_\theta$, while potentially assigning mass to regions that are improbable under $p$ (Fig.~\ref{fig:KLD}). 

One commonly estimates the KLD in the Monte Carlo approximation, by replacing the expectation in Eq.~\eqref{eq:KLD-definition} with the average over a finite set of $N$ samples from $p$. For optimization purposes, one can further omit the constant term $c = \mathbb{E}_{\mu\sim p(\mu)} \log p(\mu)$, as it is independent of $q_\theta$. The KLD
\begin{equation}\label{eq:uncond-fwd-KLD-MC}
    D_\mathrm{KL}(p||q_\theta) \approx L + c,\qquad L = - \frac{1}{N}\sum_i \log q_\theta(\mu_i),\qquad\mu_i\sim p(\mu)
\end{equation}
can thus directly be optimized by minimizing the loss $L$. 
The forward KLD loss is thus tractable whenever we can \textbf{sample the target} $\mu\sim p(\mu)$.
It is often not possible to estimate the constant $c$, as this would require access to the normalized target density $\log p(\mu)$. Therefore, the ability to \emph{optimize} the forward KLD does not imply one can \emph{evaluate} it.

\begin{figure}
    \centering
    \includegraphics[width=\linewidth]{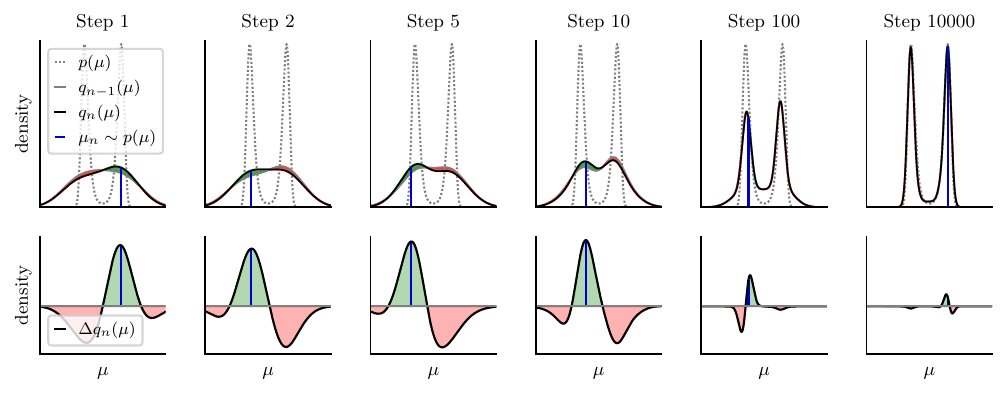}
    \caption{
    Gaussian mixture model $q_\theta$ fitted to a bimodal distribution $p$ via the forward KLD, using batch size 1. In each iteration $n$, a gradient step is taken to decrease the loss $L=-\log q_\theta(\mu_n)$, increasing the density around the training sample $\mu_n\sim p(\mu)$. The learning rate is annealed at the end of training, resulting in smaller updates.
    }
    \label{fig:uncond-fwd-KLD}
\end{figure}

Intuitively, optimization iteratively increases the density $\log q_\theta(\mu)$ at the samples $\mu_i\sim p(\mu)$ encountered in each respective batch (Fig.~\ref{fig:uncond-fwd-KLD}). The forward KLD implements no explicit mechanism to suppress probability mass in regions where $p(\mu)$ is small. Instead, probability mass is only taken away from such regions as $q_\theta(\mu)$ is normalized, so increasing the density in the vicinity of the training samples $\mu_i$ requires to decrease the density elsewhere.

The \textbf{reverse KLD} $D_\mathrm{KL}(q_\theta||p)$ on the other hand acts as a \emph{mode-seeking} (or \emph{zero-forcing}) objective. This KLD only remains finite if $p(\mu)>0$ wherever $q_\theta(\mu) > 0$, so its optimization forces $q_\theta$ to avoid placing probability mass in regions where $p$ is (near) zero. Heuristically, this ensures that all samples from $q_\theta$ are plausible under $p$, while potentially failing to capture the full support of $p$ or collapsing to a single mode in multimodal distributions (Fig.~\ref{fig:KLD}). 

In analogy to Eq.~\eqref{eq:uncond-fwd-KLD-MC}, the reverse KLD 
\begin{equation}\label{eq:uncond-rev-KLD-MC}
D_\mathrm{KL}(q_\theta||p) \approx L, \qquad L = \frac{1}{N}\sum_j \left[ \log q_\theta(\mu_j) - \log p(\mu_j) \right], \qquad \mu_j \sim q_\theta(\mu)
\end{equation}
can be optimized by minimizing the loss $L$. The reverse KLD is tractable whenever we can \textbf{evaluate the target} $p(\mu)$. For optimization purposes, it is sufficient to evaluate only the unnormalized target density $\tilde{p}(\mu)\propto p(\mu)$, in which case the loss only estimates the KLD up to an additive constant independent of $q_\theta$.

Optimizing the reverse KLD is numerically challenging due to the expectation with respect to $q$. Differentiating through the expectation $\mathbb{E}_{q_\theta(\mu)}$ usually requires to apply the reparameterization trick~\cite{kingma2013auto,rezende2014stochastic}. Optimization is further associated with an exploration problem. Training samples are generated from $q_\theta$, making it numerically difficult to expand the model support. 

In some cases, one can both sample from and evaluate the target density, such that both directions of the KLD are tractable. Then, optimization via the forward KLD is usually preferable, as it enables more stable optimization.

\subsubsection{The KLD for conditional distributions and NPE}\label{subsec:KLD-cond}

NPE extends to the conditional case, fitting a model $q_\theta(\mu|x)$ to the Bayesian posterior $p(\mu|x)$ (Eq.~\eqref{eq:bayes}). This is commonly done by minimizing the forward KLD between posterior and model, marginalizing over observations $x\sim p(x)$,
\begin{align}
    \mathbb{E}_{x\sim p(x)} D_\mathrm{KL}(p||q_\theta) 
    \label{eq:cond-KLD}
    = \mathbb{E}_{x\sim p(x)}\left[ \mathbb{E}_{\mu\sim p(\mu|x)}\log\left(\frac{p(\mu|x)}{q_\theta(\mu|x)}\right) \right]
    = \mathbb{E}_{(\mu,x)\sim p(\mu,x)}\log\left(\frac{p(\mu|x)}{q_\theta(\mu|x)}\right).
\end{align}
A naive Monte Carlo estimate of the nested expression requires posterior samples for the inner expectation, but there is usually no straightforward way to sample the posterior in the simulation-based inference paradigm. 
This is bypassed by instead sampling the joint expectation over $p(\mu,x)$, which is accessible via prior and likelihood. We can thus estimate the marginalized forward KLD via
 
\begin{equation}\label{eq:KLD-Loss-NPE}
    \mathbb{E}_{x\sim p(x)} D_\mathrm{KL}(p||q_\theta) \approx L + c,\qquad L = - \frac{1}{N}\sum_i \log q_\theta(\mu_i|x_i),\qquad\mu_i\sim p(\mu),\,x_i\sim p(x|\mu_i),
\end{equation}
where the constant $c = \mathbb{E}_{(\mu,x)\sim p(\mu,x)}\log p(\mu|x)$ is again irrelevant for the optimization of $q_\theta$.

\begin{figure}
    \centering
    \includegraphics[width=\linewidth]{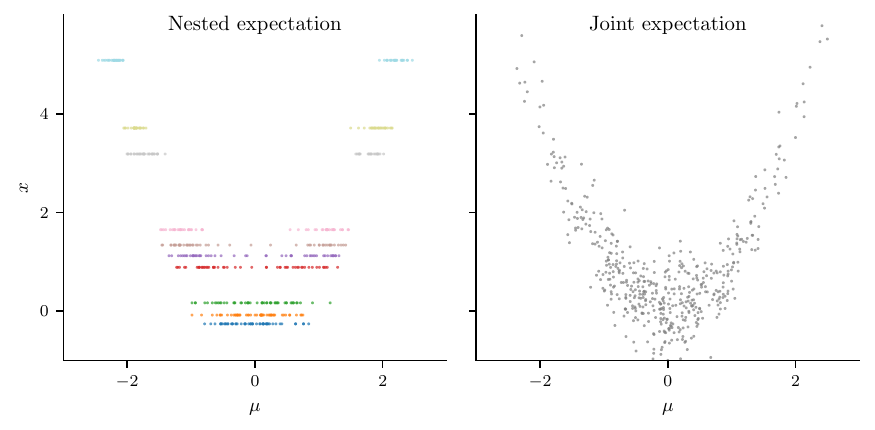}
    \caption{
    Illustration of samples used for the Monte Carlo estimation of Eq.~\eqref{eq:cond-KLD}. 
    The nested expectation can be estimated with a set of $N_x$ observations $x_i\sim p(x_i)$ and sets of $N_\mu$ posterior samples $\mu_{i,j}\sim p(\mu|x_i)$ for each of these (left; observations indicated with different colors; $N_x=10$, $N_\mu=100$).
    The joint expectation can be estimated with a set of $N$ samples $(\mu_i,x_i)\sim p(\mu,x)$ from the joint distribution (right; $N=1000$, corresponding to a nested expectation with $N_x=N$, $N_\mu=1$).
    Both ways of sampling provide valid Monte Carlo estimates, but only the joint approach is tractable in the simulation-based inference context, while also sampling the joint space more densely.
    }
    \label{fig:nested-vs-joint-expectation}
\end{figure}

\begin{figure}
    \centering
    \includegraphics[width=\linewidth]{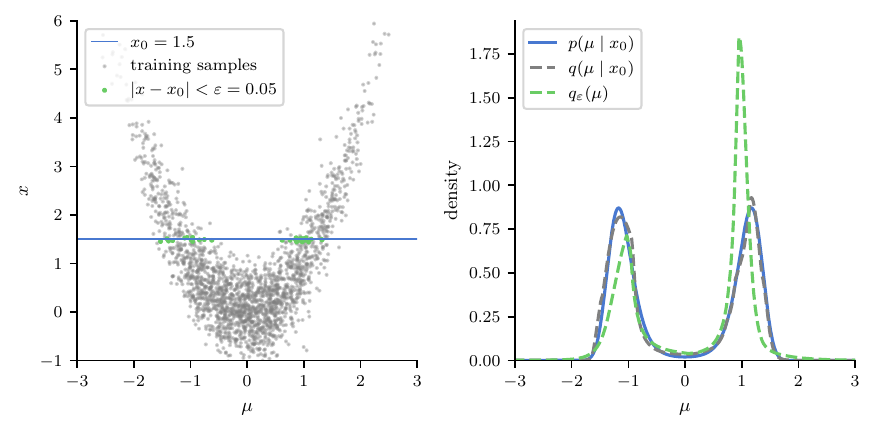}
    \caption{
    An NPE model $q_\theta(\mu|x)$ trained with all samples from the joint distribution (gray, $N=2000$) outperforms an unconditional model $q_\varepsilon(\mu)$ trained with only the subset of samples accepted in an $\varepsilon$-neighborhood of the observation of interest (green, $N=30$). The latter corresponds to Approximate Bayesian Computation and illustrates its limitation: only $30$ of the $2000$ simulations are accepted, and reducing $\varepsilon$ to improve accuracy would reject even more. NPE instead uses every simulation, extracting shared structure across observations and enabling efficient interpolation.
    }
    \label{fig:amortized-training}
\end{figure}

It may seem surprising that we can train a model $q_\theta(\mu|x)$ for the posterior without access to corresponding samples $\mu\sim p(\mu|x)$. This is achieved through amortization across observations. Sampling from the joint distribution $(\mu_i,x_i)\sim p(\mu, x)$ generates one single posterior sample $\mu_i\sim p(\mu|x_i)$ for that specific observation $x_i$. We can not choose $x_i$---this is determined by $x_i\sim p(x|\mu_i),\,\mu_i\sim p(\mu)$---and we can not generate more than one sample for this $x_i$. 
The lack of multiple samples per observation is instead compensated for by a large number of observations (Fig.~\ref{fig:nested-vs-joint-expectation}). The model then learns to interpolate between data points, leveraging shared structure among observations. Indeed, the trained model benefits also from samples that are not directly in the vicinity of the observation of interest (Fig.~\ref{fig:amortized-training}). This relies on sufficient smoothness of the likelihood in $x$.

\subsubsection{Convergence of forward KLD optimization}\label{subsec:KLD-practical}

\begin{figure}
    \centering
    \includegraphics[width=0.8\linewidth]{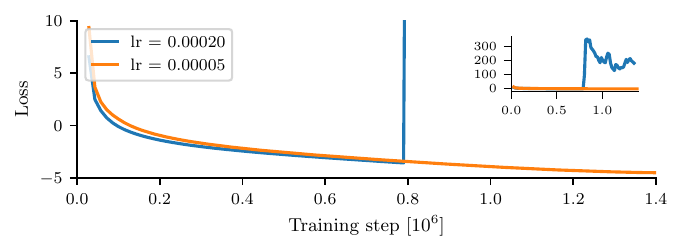}
    \caption{
    Larger learning rates typically lead to faster convergence in NPE training. However, when the learning rate is too high, volatile gradients can cause the loss to abruptly explode (blue). 
    The clean solution is to restart training from scratch with a smaller learning rate, although resuming from a model checkpoint before the corrupted training step may also work in some cases.
    Example taken from a hyperparameter sweep for NPE for gravitational-wave inference~\cite{Dax:2024mcn}.
    }
    \label{fig:KLD-lr}
\end{figure}

The forward KLD is generally a stable objective, whether for unconditional or conditional density estimation. However, it becomes volatile when the model distribution $q_\theta$ fails to fully cover the data distribution $p$. If the model assigns very low density $q_\theta(\mu_i)$ to a training example $\mu_i\sim p(\mu)$, the gradient of the loss with respect to the learnable model parameters $\theta$ diverges, $\nabla_\theta \log q_\theta(\mu) = \frac{1}{q_\theta(\mu)}\nabla_\theta q_\theta(\mu)\gg1$. 
To prevent such exploding gradients at the start of training, $q_\theta$ must be initialized to cover the entire support of $p$. A common practical approach is to normalize each dimension of the parameters $\mu$ during preprocessing, as most generative models $q_\theta$ are by default initialized to have (roughly) zero mean and unit variance.
Furthermore, stable optimization requires a carefully tuned learning rate. With a high learning rate, optimization may prematurely displace probability mass from regions of the support that have not yet been sufficiently sampled. High learning rates also amplify the impact of exploding gradients when outlier samples with very low density $q_\theta(\mu_i)$ are encountered. 
As a rule of thumb, the initial learning rate should be chosen as large as possible while still avoiding exploding gradients (Fig.~\ref{fig:KLD-lr}). 
Towards the end of training, it is further useful to anneal the learning rate or increase the batch size to improve convergence.
Training stability can also be improved with gradient clipping~\cite{pascanu2013difficulty}, in particular in low-data regimes. 

The loss estimates the KLD up to a constant, $D_\text{KL}=L+c$. This constant, and therefore also the optimal loss $L_\text{opt}=-c$, is usually unknown. We can therefore only assess the loss in relative terms, for example to compare different models. Note that $c$ depends on the prior, the likelihood and the normalization of $\mu$, so comparison between models with different choices for any of these is usually not straightforward (although in some cases $\Delta c$ can be computed analytically).

As NPE generates the training data from the joint distribution, it trains by construction with ground-truth posterior samples. The parameter space is thus explored through a global mechanism, accurately representing distributional modes. This is an inherent advantage compared to inference methods that rely on local exploration, such as MCMC. Moreover, due to the mass covering objective, failure to capture all distributional modes in the training data is flagged with a diverging loss.

Global sampling comes at the cost of losing control over the weight assigned to different observations in the marginalized KLD $\mathbb{E}_{x\sim w(x)} D_\mathrm{KL}(p||q_\theta)$. This is because the joint expectation in Eq.~\eqref{eq:cond-KLD} can only be sampled through likelihood and prior when weighting $x$ with the evidence, 
\begin{equation}\label{eq:w-mu}  
    w(x)=\int\d\mu\, p(\mu)\,p(x|\mu)=p(x). 
\end{equation}
In theory, any $w(x)$ (with sufficient support) yields the same optimal $q_\theta(\mu|x)$. In practice however, $w(x)$ defines the training distribution, and therefore determines how well the model is trained for different observations. 
The evidence is a natural weighting, prioritizing observations that are likely under the assumed prior and likelihood, focusing training capacity on typical events.
However, sometimes rare events are especially scientifically interesting, and weighting with the evidence suppresses these in training. 
The lack of flexibility with respect to $w(x)$ also impedes finetuning NPE for specific observations, which instead requires specialized frameworks such as multi-round inference (Sec.~\ref{subsec:sequential-inference}) or a likelihood-based loss~\cite{schmitt2024leveraging,mishra2026robust}. 

\subsubsection{Sequential inference}\label{subsec:sequential-inference}
Sometimes one is only interested in a single observation $x_0$. Fully amortized NPE or NLE can then be inefficient, as only a small fraction of the training distribution $w(x)$ may be informative about the target $p(\mu|x_0)$ or $p(x_0|\mu)$.
Sequential NPE~\cite{papamakarios2016fast,lueckmann2017flexible,greenberg2019automatic,deistler2022truncated} and NLE~\cite{papamakarios2019sequential} iteratively enhance the focus on $x_0$ across multiple rounds. 
In the first round, the training data is generated as usual, by sampling parameters from prior $\mu_i\sim p(\mu)$.
In each following round $n$, the prior is then replaced by the posterior estimate from the previous round, $\mu_i\sim q_\theta^{n-1}(\mu|x_0)$. 
This iteratively constrains the parameter space to the region relevant for the observation $x_0$.

Sequential NPE requires an importance weighted loss to ensure that the network estimates the posterior with respect to the desired physical prior. The weights may have a high variance in practice and need to be carefully regularized~\cite{papamakarios2016fast,lueckmann2017flexible,greenberg2019automatic,deistler2022truncated,lueckmann2021benchmarking}. For NLE, such importance weights are not necessary.

While sequential methods are efficient for individual observations, amortized inference is usually better suited in contexts where inference needs to be repeated for many different observations or where fast analysis is critical.

\subsection{Ratio estimation}\label{subsec:ml-classifiers}

\begin{figure}
    \centering
    \includegraphics[width=0.49\linewidth]{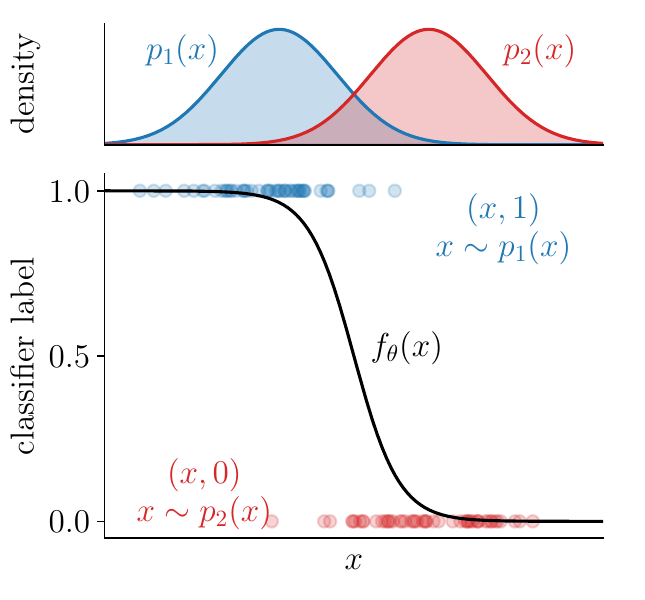}
    \includegraphics[width=0.49\linewidth]{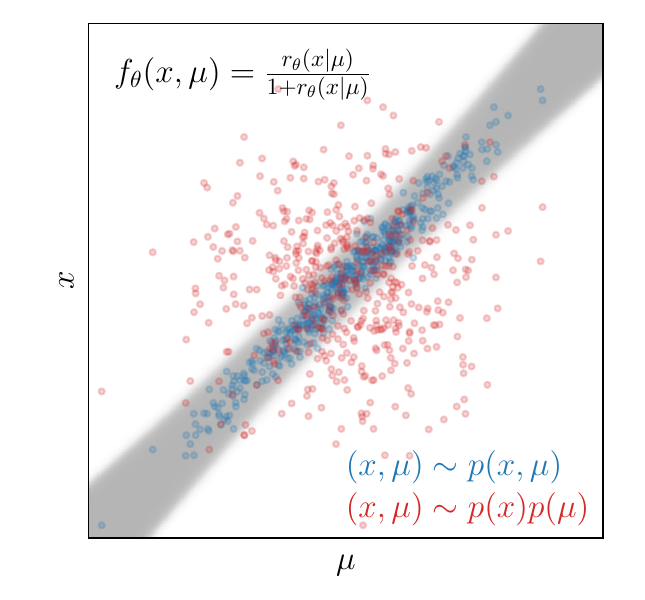}
    \caption{
    Left panel: illustration of likelihood ratio trick, where a classifier $f_\theta(x)$ is trained between samples from two distributions $p_1(x)$ and $p_2(x)$. Right panel: application of the likelihood ratio trick to learn the likelihood-to-evidence ratio $r(x|\mu)$. The blue and red dots are samples from the joint distribution $p(x,\mu)$ and the product of the marginals $p(x)p(\mu)$. The grey shading shows the learned classifier label. 
    }
    \label{fig:nre}
\end{figure}

So far, we have discussed how generative networks can be used to directly learn (conditional) probability distributions, enabling cheap sampling from them. Instead of learning distributions directly, it is also possible to learn ratios between distributions using classifier networks. This allows for much more flexibility in choosing the network architecture than very constrained models like normalizing flows, which must be invertible. However, to perform Bayesian inference by sampling from the posterior, the network has to be combined with a sampling algorithm like MCMC, resulting in the loss of full amortization. On the other hand, frequentist inference in the limit of many observations remains amortized, as discussed in more detail in Sec.~\ref{subsec:combine_obs}.

\subsubsection{Likelihood-ratio trick}\label{subsec:lrtrick}

Consider two probability distributions $p_1(x)$ and $p_2(x)$ with a ratio $r(x)$ that we want to estimate using a neural network $r_\theta(x)$,
\begin{equation}
    r_\theta(x) \approx r(x) = \frac{p_1(x)}{p_2(x)} \;.
\end{equation}
Often, evaluating the probability densities $p_1$ and $p_2$ directly is not tractable, but we have access to samples from these distributions. We are still able to extract the ratio $r(x)$ by training a classifier network using a binary cross-entropy loss, where $f_\theta(x)$ denotes the classifier output,
\begin{equation}
\begin{split}
    L &= - \mathbb{E}_{x \sim p_1(x)}\big[ \log f_\theta(x) \big] - \mathbb{E}_{x \sim p_2(x)}\big[ \log (1 - f_\theta(x)) \big] \\
    &= - \int\big[ p_1(x) \log f_\theta(x) + p_2(x) \log (1 - f_\theta(x)) \big] \,\d x \;.
\end{split}
\end{equation}
We can extract the function $f^*(x)$ that minimizes the loss function by computing the functional derivative,
\begin{equation}
    \label{eq:classifier_loss}
    \frac{\delta L}{\delta f} = - \frac{p_1(x)}{f(x)} + \frac{p_2(x)}{1 - f(x)}
    \overset{!}{=} 0
    \qquad\implies\qquad
    f^*(x) = \frac{p_1(x)}{p_1(x) + p_2(x)}
    \;.
\end{equation}
The left panel of Fig.~\ref{fig:nre} illustrates the function learned by a classifier trained with this loss function. Rewriting Eq.\eqref{eq:classifier_loss} in terms of the likelihood ratio yields
\begin{equation}
    r(x) = \frac{p_1(x)}{p_2(x)} = \frac{f^*(x)}{1 - f^*(x)} \;.
\end{equation}
Consequently, given a sufficiently expressive network $f_\theta(x)$, the learned ratio $r_\theta(x)$ approximates the true ratio $r(x)$~\cite{goodfellow2014generative,Cranmer:2015bka,sugiyama2012density,hastie2009elements,hermans2020likelihood}. This is often referred to as the \textbf{likelihood-ratio trick}~\cite{miller2022contrastive,Rizvi:2023mws}.

\subsubsection{Neural Ratio Estimation (NRE)}\label{subsec:nre}

One possible option is to learn the likelihood-to-evidence ratio $r(x|\mu)$~\cite{hermans2020likelihood,durkan2020contrastive,miller2022contrastive}. It can equivalently be written as the ratio between the joint distribution $p(x,\mu)$ and the product of the marginal distributions $p(x)$ and $p(\mu)$, or as the ratio of the posterior to the prior,
\begin{equation}
    r(x|\mu) = \frac{p(x|\mu)}{p(x)} = \frac{p(x,\mu)}{p(x)p(\mu)} = \frac{p(\mu|x)}{p(\mu)} \;.
\end{equation}
This ratio can be obtained by training a classifier on samples from the joint distribution and a shuffled version of the same samples to remove the pairing between the parameters $\mu$ and the observations $x$, as illustrated in the right panel of Fig.~\ref{fig:nre}.

We can then use the estimated likelihood-to-evidence ratio both for Bayesian and frequentist inference. In the Bayesian case, we have to multiply it by the prior, yielding the posterior,
\begin{equation}
    p(\mu|x) = p(\mu)\, r(x|\mu) \;.
\end{equation}
However, unlike neural posterior estimation, we cannot directly sample from the posterior. Instead, we have to combine the estimated ratio with a sampling procedure such as Markov Chain Monte Carlo (MCMC), nested sampling, or importance sampling. While this incurs a higher computational cost than directly sampling from the posterior, it still means we do not have to run the costly simulation once the network is trained. Furthermore, for frequentist inference, we need the likelihood ratio between the likelihoods for two hypotheses $\mu_1$ and $\mu_2$, which can be easily computed from the likelihood-to-evidence ratio,
\begin{equation}
    \label{eq:nre_likeli_ratio}
    r(x|\mu_1, \mu_2) = \frac{p(x|\mu_1)}{p(x|\mu_2)} = \frac{r(x|\mu_1)}{r(x|\mu_2)} \; .
\end{equation}

Despite its flexibility, training a classifier between the joint distribution and the product of the marginals can be challenging in some cases. For instance, if the joint distribution is narrow, most points from the distribution $p(\mu) p(x)$ will be in regions where $p(\mu, x)$ is small, leading to noisy gradients during training and, consequently, poor estimates of $r(x|\mu)$. This problem can be avoided in a frequentist setup by learning the likelihood ratio from Eq.\eqref{eq:nre_likeli_ratio} directly~\cite{Cranmer:2015bka}, either by assuming a fixed null hypothesis $\mu_2 = \mu_0$ or by training a classifier with two inputs for $\mu_1$ and $\mu_2$. However, even in this setup instability can arise when the supports of the distributions only have a small overlap. %

\subsection{Incorporating likelihood information}
\label{subsec:ml-likelihood}

Simulation-based inference enables inference when the likelihood is intractable and accelerates it even when the likelihood is known. In the following, we discuss how density estimation can be augmented with importance sampling for asymptotically exact inference in the case of an analytically known likelihood in Sec.~\ref{subsec:npe-is}, and how a known likelihood $p(z|\theta)$ at a latent simulation step can be exploited during training to reduce the variance of gradients and improve convergence in Sec.~\ref{sec:latent_likelihood}.

\subsubsection{Importance sampling}\label{subsec:npe-is}

\begin{figure}
    \centering
    \includegraphics[width=0.49\linewidth]{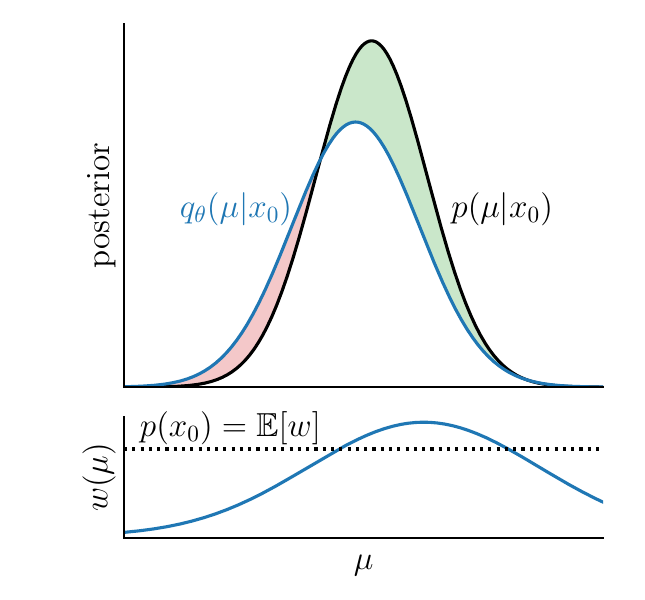}
    \includegraphics[width=0.49\linewidth]{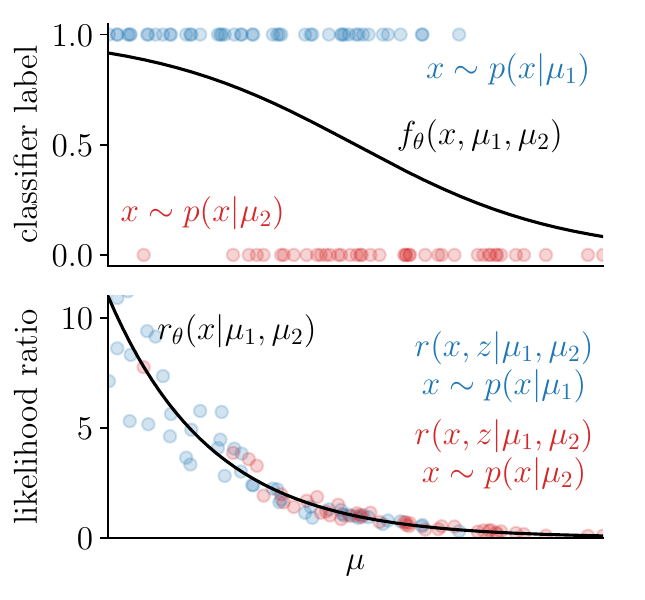}
    \caption{
    Left: importance sampling reweights samples from the NPE proposal (blue) to the ground-truth posterior (black), suppressing overestimated (red) and amplifying underestimated (green) regions. The normalization of the weights yields an unbiased Monte Carlo estimated of the Bayesian evidence (lower panel).
    Right: illustration of how learning the likelihood ratio between two parameters $\mu_1$ and $\mu_2$ can be turned from a classification task (upper panel) into a regression task (lower panel) by computing the joint ratio $r(x,z|\mu_1,\mu_2)$ using latent likelihood knowledge.
    }
    \label{fig:incorporate_likeli}
\end{figure}

While simulation-based training relies on sampling the likelihood, it is often also possible to evaluate its density. The posterior $p(\mu|x)\propto p(x|\mu)p(\mu)$ can then be evaluated up to the normalization $p(x)$, and one can augment the trained NPE network with importance sampling (NPE-IS~\cite{Dax:2022pxd}). This generates a set of weighted samples
\begin{equation}
    \label{eq:is_weights}
    \mu_i\sim q_\theta(\mu|x), \quad
    w_i = \frac{p(x|\mu_i)p(\mu_i)}{q_\theta(\mu_i|x)}.
\end{equation}
The importance weights account for deviations between the NPE proposal and the target density, with high weights amplifying underrepresented and low weights suppressing overrepresented regions (Fig.~\ref{fig:incorporate_likeli}). A set of $N$ weighted samples $(\mu_i,w_i)$ has the same statistical power as\looseness=-1
\begin{equation}
    N_\mathrm{eff} = \frac{\left(\sum_{i=1}^N w_i\right)^2}{\sum_{i=1}^N w_i^2}
\end{equation}
unweighted samples from the ground-truth posterior. Hence, $N_\mathrm{eff}$ is referred to as the effective sample size~\cite{kish1965survey,kish1992weighting}. The sample efficiency $\varepsilon=N_\mathrm{eff} / N \in (0, 1]$ arises as a performance measure. A perfect NPE network yields constant weights and therefore $N_\mathrm{eff} = N$ and $\varepsilon=1$. The evidence integral can be computed from the normalization of the weights
\begin{equation}
    p(x) = \int\frac{p(x|\mu)\, p(\mu)}{q_\theta(\mu|x)} q_\theta(\mu|x)\, \d\mu
    = \mathbb{E}_{\mu \sim q_\theta(\mu|x)}\big[ w(\mu) \big]
    \approx \frac{1}{N}\sum_{i=1}^{N}w_i,
\end{equation}
The statistical uncertainty of the log evidence, $\sigma_{\log p(x)} = \sqrt{(1-\varepsilon)/(N\cdot\varepsilon)}$, scales with $1/\sqrt{N}$, enabling precise estimates~\cite{Dax:2022pxd}. 

Importance sampling has two main pitfalls. First, if the proposal $q_\theta(\mu|x)$ does not cover the full support of $p(\mu|x)$---for example, by missing an entire distributional mode---importance sampling cannot recover the missing probability mass since samples are drawn exclusively from $q_\theta$. This results in an underestimate of the evidence and biased posterior samples. Fortunately, NPE models should usually cover the full posterior support, as they are trained with the mass-covering forward KLD.
Second, a poor or light-tailed approximation $q_\theta$ may result in low sample efficiency, requiring prohibitively many samples to achieve a given $N_\mathrm{eff}$. Importance sampling is unforgiving, as the efficiency is sensitive to any deviation between proposal and target, even in high-dimensional correlations. Importance sampling thus requires the initial NPE model to be already a good approximation of the posterior. Conversely, a low sample efficiency does not imply that $q_\theta$ is a poor approximation of the posterior, as relevant features such as low-dimensional marginals may still be adequately captured.

When the likelihood is tractable and the initial NPE model is sufficiently accurate, NPE-IS is powerful: it turns an approximate posterior into an asymptotically exact one, while providing a strong performance diagnostic for validation (Sec.~\ref{subsec:compare-high-dim}) and an estimate of the Bayesian evidence. This is particularly useful in scientific applications with high reliability requirements.

\subsubsection{Known latent likelihood}
\label{sec:latent_likelihood}

In some cases, we do not know the full likelihood of the simulation but have access to a tractable likelihood for an intermediate simulation step (see Eq.\eqref{eq:forward_model_latent}). This is a typical situation in particle physics, where we have a whole chain of simulation tools, starting from the computation of scattering amplitudes using quantum field theory. Often, this is the only simulation step influenced by the parameters of interest, and it has a tractable likelihood. While it is not possible to perform importance sampling in such cases, we can still leverage latent likelihood information during training~\cite{Brehmer:2018hga,Brehmer:2018eca}. If the likelihood of the observed data $x$ given the latent space $z$ does not depend on the model parameters $\mu$, the joint likelihood ratio $r(x, z | \mu_0, \mu_1)$ is equal to the known likelihood ratio in the latent space,
\begin{equation}
    r(x, z | \mu_0, \mu_1)
    = \frac{p(x,z|\mu_0)}{p(x,z|\mu_1)}
    = \frac{p(x|z) \: p(z|\mu_0)}{p(x|z) \: p(z|\mu_1)}
    = \frac{p(z|\mu_0)}{p(z|\mu_1)} \; .
\end{equation}
While both $x$ and $z$ are known for the simulated training points, at inference time, only $x$ is available. Using variational calculus, it can be shown that a loss function of the form
\begin{equation}
    \label{eq:joint_ratio_loss}
    L = \mathbb{E}_{(x,z) \sim p(x, z | \mu_1)} \left[(r(x, z | \mu_0, \mu_1) - r_\theta(x | \mu_0, \mu_1))^2 \right]
\end{equation}
is minimized by the function
\begin{equation}
    r_*(x | \mu_0, \mu_1) = \int p(z|x, \mu_1) \, r(x, z | \mu_0, \mu_1)\, \d z \;.
\end{equation}
Inserting the definition of the joint likelihood ratio yields the likelihood ratio for $x$,
\begin{equation}
\label{eq:ratio_cancellation}
\begin{split}
    r_*(x | \mu_0, \mu_1)
    &= \int p(z|x, \mu_1) \, \frac{p(x,z|\mu_0)}{p(x,z|\mu_1)} \,\d z \\
    &= \int p(z|x, \mu_1) \, \frac{p(z|x,\mu_0) \, p(x|\mu_0)}{p(z|x,\mu_1)\, p(x|\mu_1)}\,\d z
    = \frac{p(x|\mu_0)}{p(x|\mu_1)} \;.
\end{split}
\end{equation}
Estimating the likelihood ratio is, therefore, turned from a classification task into a regression task. If the joint likelihood ratio is a good approximation of the ratio in data space, this leads to less noisy gradients during training compared to a classifier training, as illustrated in the right panel of Fig.~\ref{fig:incorporate_likeli}. The loss function from Eq.\eqref{eq:joint_ratio_loss} can be easily extended to training where both $\mu_0$ and $\mu_1$ are varied.

We can extract even more information from the latent likelihood if its gradients with respect to $\mu$ are tractable. This provides access to the joint score,
\begin{equation}
    t(x, z | \mu) = \nabla_\mu \log p(x, z | \mu)
    = \frac{\nabla_\mu p(x, z | \mu)}{p(x, z | \mu)} \;.
\end{equation}
We again use the argument from Eq.\eqref{eq:ratio_cancellation} to write it as a function of the latent space likelihood $p(z|\mu)$,
\begin{equation}
    t(x, z | \mu) = \frac{\nabla_\mu p(z | \mu)}{p(z | \mu)} \;.
\end{equation}
Using the same trick as for the likelihood ratio in Eq.\eqref{eq:joint_ratio_loss}, the mean of the joint score with respect to the latent space $z$ gives the score $t(x|\mu)$. This lets us use a regression loss between the joint score and the score of the learned function in addition to the regression or classification loss for ratio estimation, $L_\text{NRE}$, with a hyperparameter $\lambda$ to balance the terms,
\begin{equation}
    L_\text{score} = L_\text{NRE} + \lambda\, \mathbb{E}_{(x,z) \sim p(x, z | \mu)} \left[ (t(x, z | \mu_0) - \nabla_\mu \log r_\theta(x|\mu_0, \mu_1))^2 \right] \;.
\end{equation}
Adding a score-matching loss is not limited to ratio estimation, and can also be applied to density estimation with loss $L_\text{DE}$ as long as the learned density is tractable and differentiable,
\begin{equation}
    L_\text{score} = L_\text{DE} + \lambda\, \mathbb{E}_{(x,z) \sim p(x, z | \mu_1)} \left[ (t(x, z | \mu) - \nabla_\mu \log q_\theta(x|\mu))^2 \right] \;.
\end{equation}
This loss function is also conceptually related to the training of score-based density estimators, with the difference that for such models the learned score is directly parameterized by a neural network, $t_\theta(\mu|x,t)$, and the target score is defined for a $t$-dependent trajectory from random noise to the target distribution, $p(\mu(t)|\mu(0), x, t)$.

\subsection{Distribution-level inference}\label{subsec:ml-emp-bayes}

Previously, we have shown how density estimation and ratio estimation can be used for parameter inference, both in a Bayesian and frequentist setting. These methods, however, are not limited to performing inference for individual data points and can also be applied at the level of distributions, i.e., for Empirical Bayes or unfolding tasks as defined in Eq.\eqref{eq:dist_level_inf}. In the following, we show how this can be achieved either by an iterative procedure or by directly maximizing the likelihood in Eq.\eqref{eq:dist_level_opt}.

\subsubsection{Iterative methods}

\begin{figure}
    \centering
    \includegraphics[width=0.49\linewidth]{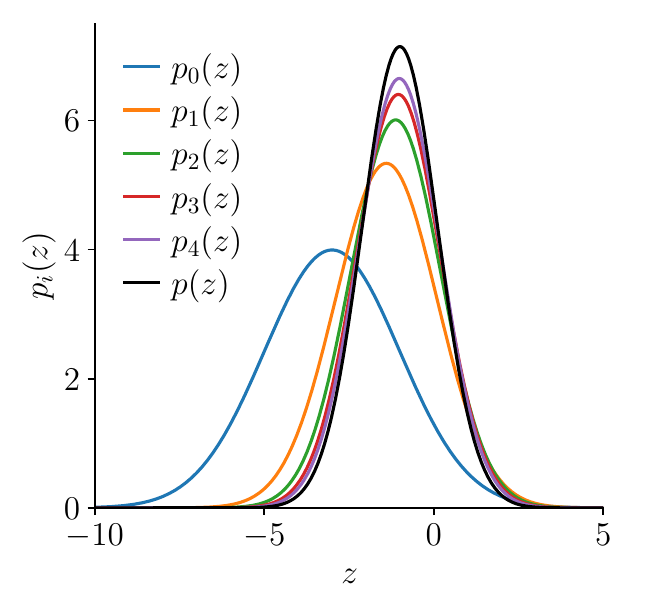}
    \includegraphics[width=0.49\linewidth]{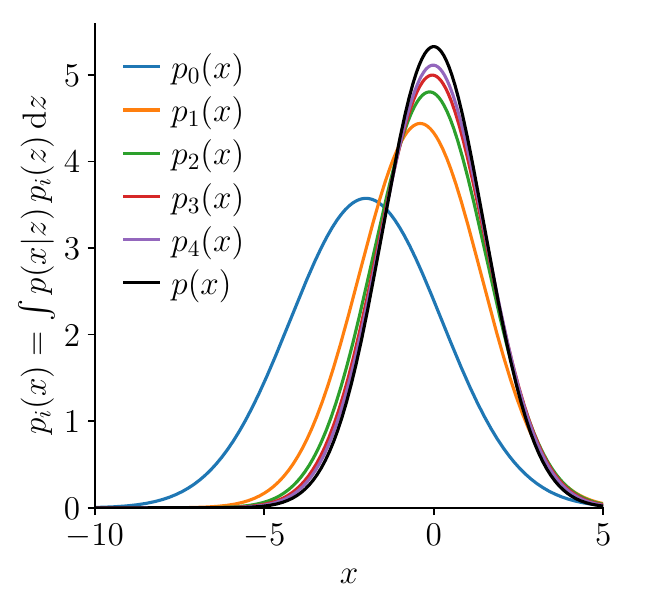}
    \caption{
    Illustration of the iterative procedure to find the distribution $p(z)$ that recovers the data distribution $p(x)=\mathcal{N}(x;\mu=0,\sigma=1.5)$ after a Gaussian smearing $p(x|z)=\mathcal{N}(x-z;\mu=1,\sigma=1)$ is applied. The left panel shows the approximations $p_i(z)$ starting from the initial distribution $p_0(z)$ compared to the target $p(z)$, and the right panel shows the distributions after applying the forward process compared to the data distribution $p(x)$.
    }
    \label{fig:iterative}
\end{figure}

Iterative methods work by assuming an initial guess, or prior, $p_0(z)$, and then iteratively reducing the dependence of the result on the prior to find a result that fulfills Eq.\eqref{eq:dist_level_inf}. This can be achieved using the expectation-maximization algorithm~\cite{dempster1977maximum}:
\begin{enumerate}
    \item Define the posterior based on the previous assumption for the prior, $p_{i-1}(z)$, 
    \begin{equation}
        \label{eq:iter_unf_posterior}
        p_{i-1}(z|x) = \frac{p(x|z)\, p_{i-1}(z)}{\int p(x|z')\, p_{i-1}(z')\,\d z'} \;.
    \end{equation}
    \item Update the prior by applying the posterior to the observed data,
    \begin{equation}
        \label{eq:iter_unf_update}
        p_i(z) = \int p(x)\, p_{i-1}(z|x) \,\d x \;.
    \end{equation}
\end{enumerate}
Following Eq.\eqref{eq:dist_level_inf}, our target is the distribution $p_*(z)$ fulfilling
\begin{equation}
    p(x) = \int p(x|z)\, p_*(z)\, \d z \;.
\end{equation}
By plugging $p_*(z)$ into Eqs.\eqref{eq:iter_unf_posterior} and~\eqref{eq:iter_unf_update}, we can see that the target is a fixed point of the iterative procedure above,
\begin{equation}
\begin{alignedat}{2}
    \int p(x)\, p_*(z|x)\, \d x &= \int p(x) \frac{p(x|z)\, p_*(z)}{\int p(x|z')\, p_*(z')\, \d z'} \, \d x && \\
    &= \int p(x) \frac{p(x|z)\, p_*(z)}{p(x)}\d x &&= p_*(z) \;.
\end{alignedat}
\end{equation}
Each iteration $p_i(z)$ reduces the dependence on the prior assumption $p_0(z)$, and it can be shown that the limit of this iteration is indeed the fixed point $p_*(z)$~\cite{Andreassen:2019cjw,dempster1977maximum,wu1983convergence,vardi1985statistical,kuusela2012statistical}. Fig.~\ref{fig:iterative} shows this iterative process for an example distribution. Note that this procedure is related to sequential inference as described in Sec.~\ref{subsec:sequential-inference}, with the difference that in sequential inference the posterior for a single observation $x_0$ is used to update the prior. In contrast, in distribution-level inference the posterior is integrated out over the distribution of observed data before updating the prior.

Having reduced the task of distribution-level inference to repeated application of Bayesian inference, we can again use neural posterior or ratio estimation to perform the iteration described above~\cite{Huetsch:2024quz}. In both cases, we start from an initial assumption $p_0(z) \equiv p_\text{sim}(z)$ for the target distribution. We then use the simulator to obtain paired samples,
\begin{equation}
    (x,z) \sim p_\text{sim}(x,z) = p(x|z)\, p_\text{sim}(z) \;.
\end{equation}
In the \textsc{OmniFold} method based on ratio estimation~\cite{Andreassen:2019cjw}, we first train a classifier between simulations and the observed data, which yields their ratio,
\begin{equation}
    w_\theta(x) \approx \frac{p(x)}{p_\text{sim}(x)} \;.
\end{equation}
Next, the paired simulated points $(x,z)$ allow us to pull back these weights to $z$-space. The resulting distribution is
\begin{equation}
    p_1(z) = \int p_\text{sim}(x, z) \, w_\theta(x)\, \d x = \int  p_\text{sim}(z | x) \, p(x)\, \d x \;.
\end{equation}
This achieves the update step from Eq.\eqref{eq:iter_unf_update}.

Alternatively, we can choose generative networks for the update by first estimating the posterior $q_\theta(z|x)$~\cite{Bellagente:2020piv,Backes:2022sph}, for instance using a normalizing flow or flow matching model, and then use it to sample from the updated distribution from Eq.\eqref{eq:iter_unf_update},
\begin{equation}
    z \sim q_\theta(z|x) \qquad\mwith\qquad x \sim p(x) \;.
\end{equation}

In both cases, we can obtain a function for reweighting the initial distribution $p_\text{sim}(z)$ to the updated one as a second step by training a classifier,
\begin{equation}
    v_\psi(z) \approx \frac{p_1(z)}{p_\text{sim}(z)} \;.
\end{equation}
This function then allows us to reweight the training data and perform further update steps. Usually, only a small number of iterations is performed, as they can amplify statistical fluctuations in the data, and it is often enough to sufficiently reduce the dependence on the prior $p_0(z)$.

\subsubsection{Direct optimization}

As an alternative to the iterative procedure, the distribution $q_\theta(z)$ can also be optimized directly. The training objective then is to minimize the difference between the distribution of the observed data and the distribution recovered from applying the forward mapping to the learned latent distribution,
\begin{equation}
    \label{eq:neural_empirical_bayes}
    D_\mathrm{KL}[p(x) || q_\theta(x)] = - \int p(x)\, \log \int p(x|z)\, q_\theta(z)\,\d z\,\d x + \text{const} \;.
\end{equation}
Methods based on this approach are known as Neural Empirical Bayes methods~\cite{vandegar2021neural,Butter:2025via}. The two main challenges compared to the iterative approach are that these methods require a surrogate for the forward-mapping density, and that Eq.\eqref{eq:neural_empirical_bayes} contains an integral over the logarithm that must be evaluated for each training sample. This makes the training task in Neural Empirical Bayes models more complex and computationally more costly than the iterative methods described earlier.

\subsection{Data modality}\label{subsec:ml-data-modality}

To extract as much information as possible from the observed data, it is crucial to choose an inference setup that accounts for the structure of the data. In the following, we will describe how various data modalities can be tackled with embedding networks, and then discuss the important special case of combining independent observations.

\subsubsection{Embedding networks}

Many conditional generative architectures, such as normalizing flows, can only accept fixed-size vectors as their conditional inputs. To estimate the posterior for more complex data modalities, they are combined with an \textbf{embedding network} $h_\psi(x)$ that reduces them to a vector of summary statistics. These can then be trained jointly with the generative network itself. The log-likelihood loss then reads
\begin{equation}
    L = -\mathbb{E}_{(\mu, x) \sim p(\mu, x)}\left[ \log q_\theta(\mu | h_\psi(x)) \right] \;.
\end{equation}
This approach is not limited to posterior estimation with a log-likelihood loss. Similarly, joint training of the embedding network is possible for other types of generative network training, such as flow matching and neural ratio estimation. However, joint training is not possible in neural likelihood estimation, as the summary statistics act on the target $x$, so the distribution that the network has to learn changes as the summary statistics change. In contrast, in posterior estimation the summary statistics only affect the condition, meaning that sub-optimal summary statistics only mean that the posterior is conditioned on less information.

The choice of the summary network architecture depends on the data modality. Set-like data can be processed using DeepSets~\cite{zaheer2017deep} or other permutation-equivariant architectures, such as Transformers~\cite{vaswani2017attention}, followed by a pooling layer. Time series can be processed using recurrent neural networks (RNNs) or long short-term memory (LSTM) networks~\cite{mienye2024recurrent}. Graph-like structures in the data can be exploited using graph neural networks~\cite{zhou2020graph}. Suitable network architectures for images include convolutional neural networks and vision transformers~\cite{dosovitskiy2021an}, again followed by appropriate pooling or flattening stages.

The learned summary statistics $h_\psi(x)$ can also be used independently of the posterior or ratio estimate. In combination with a regularizing loss term, they have been proposed as a way to interpret the importance of input features in network predictions and to detect model misspecification~\cite{schmitt2023detecting}.

\subsubsection{Combining observations}
\label{subsec:combine_obs}

Consider an inference task in which many independent data points describe the same parameters. For instance, in particle physics, we detect large numbers of collider events that are all governed by the same underlying model parameters. While this can be tackled with an embedding network for set-like data, the approach quickly becomes limiting. The network only processes sets within the size limits seen during training; otherwise, it starts extrapolating. For sets of thousands or even millions of events, training becomes prohibitively expensive. This can be solved by performing inference for a single event instead, followed by a procedure to combine the results.

The combined likelihood for multiple i.i.d. data points ${x_1, \ldots, x_n}$ is given in Eq.\eqref{eq:combine_likelihoods}. Using Bayes’ theorem, the posterior is given by
\begin{equation}
    \label{eq:combine_posteriors}
    p(\mu|x_1, \ldots, x_n) \propto \frac{p(\mu|x_1) \cdots p(\mu|x_n)}{p(\mu)^{n-1}} \; .
\end{equation}
Consequently, we can perform posterior estimation at the level of single observations. However, because we can only sample from the posterior for individual observations $x_i$, not from the combined posterior, the full amortization is lost. Instead, we have to combine the posterior estimate with a sampling method, such as MCMC, to obtain posterior samples. Furthermore, as the prior implicitly appears to the power $n$ in the per-event posteriors Eq.\eqref{eq:combine_posteriors} and has to be divided out, small errors in learning the prior from the training data are amplified.

For neural likelihood and ratio estimation, the inference results for independent observations can be combined by multiplying the inferred likelihoods or likelihood ratios,
\begin{equation}
    \label{eq:combine_ratios}
    r(x_1, \ldots, x_n|\mu_1, \mu_2) = r(x_1 | \mu_1, \mu_2) \cdots r(x_n | \mu_1, \mu_2) \;.
\end{equation}
This property becomes especially useful in frequentist inference, where, in the limit of many observations, the log-likelihood ratio follows a $\chi^2$ distribution. We can then extract confidence intervals cheaply, and the inference is fully amortized. This enables inference even for millions of data points at low computational cost. In addition, machine-learning based methods have been proposed to make frequentist inference amortized even outside of the asymptotic limit~\cite{dalmasso2024likelihood}. Note that combining observations becomes more complicated in the presence of nuisance parameters. The multiplicative property of likelihoods and their ratios in Eqs.\eqref{eq:combine_likelihoods} and~\eqref{eq:combine_ratios} does not hold in general for profile likelihoods. Instead, the network would have to be trained conditionally on both the parameters of interest and the nuisance parameters, and profiling would have to be performed at the level of the combined likelihood. In practice, combining large numbers of nuisance parameters and a profile likelihood approach with neural simulation-based inference has been studied for particle physics applications~\cite{ATLAS:2025clx}. However, a fully conditional treatment of the nuisance parameters was not computationally feasible. Instead, a factorized approach was used, where a correction to the likelihood ratio was learned per nuisance parameter using a training sample with nominal values of the parameter, and two samples with values above and below the nominal one.

\clearpage

\section{Validation of simulation-based inference}\label{sec:verification}
All predictions in deep simulation-based inference are subject to neural network approximations. It is thus critical to thoroughly validate an inference network before deployment in scientific data analysis. However, validation is an inherently difficult task. Comparing two high-dimensional distributions is challenging in itself even if both are tractable, but in simulation-based inference the target is usually not even directly accessible.
This limits the set of tractable tests. 
Almost all existing validation methods are \emph{necessary} but not \emph{sufficient} conditions for accurate inference.
Furthermore, validation requirements depend on the scientific downstream analyses. For instance, there is no universally correct choice for how to aggregate results across observations (is high accuracy for most observations but poor accuracy for some preferable over consistent medium accuracy?) or which metric best captures the scientific requirements (what types of inaccuracies are acceptable?).

It is therefore useful to apply a combination of multiple validation methods, depending on the scientific requirements and on which resources are available---including alternative inference methods as a reference (Sec.~\ref{subsec:comparison-to-ref}) or access to a tractable likelihood (Sec.~\ref{subsec:compare-high-dim}). Classifier tests (Sec.~\ref{subsec:compare-high-dim}), calibration tests (Sec.~\ref{subsec:calibration}) and posterior predictive checks (Sec.~\ref{subsec:posterior-predictive-checks}) are almost always applicable and serve as a useful first sanity check. 
Combining multiple such tests yields increased evidence for reliable inference, but provides no formal guarantees.
We emphasize, however, that this is true for any inference method. Convergence guarantees of conventional techniques such as MCMC or nested sampling also only hold asymptotically, and their diagnostics are only necessary but not sufficient tests of accurate inference.

\subsection{Training convergence}
As with any deep learning approach, the loss is a first useful diagnostic to assess neural network convergence. It is typically difficult to interpret directly, as it marginalizes over observations and since its optimal value is unknown. For example, the NPE/NLE loss in Eq.~\eqref{eq:KLD-Loss-NPE} estimates the marginalized KL divergence only up to an unknown constant $c$ (with a tractable likelihood, one could in principle estimate $c$ through importance sampling, but this estimate would in most cases be extremely expensive and have a high variance).

Nevertheless, the loss can be used to compare different inference networks trained with identical datasets and training objectives. Moreover, it is critical to monitor potential overfitting by comparing training and validation loss. Unlike in standard supervised learning, where overfitting primarily degrades performance on unseen data, in simulation-based inference an overfitted network also fails for the training observations themselves: since each $x_i$ (most likely) appears only once in the training set paired with a single $\mu_i$, memorization causes the network estimate to collapse to a single point, neglecting other parameter values that are also compatible with $x_i$. This is a direct consequence of the simulation-based setup, where each $(\mu_i,x_i)\sim p(\mu_i,x_i)$ pair is a single sample from the joint distribution and thus conveys only partial information about the posterior.

\subsection{Comparison to reference results}\label{subsec:comparison-to-ref}

In some applications, trusted conventional inference methods are available, for example, based on MCMC or nested sampling. These can serve as a reference to validate simulation-based inference results. A key advantage over calibration-based diagnostics (Sec.~\ref{subsec:calibration}) is that one can freely choose the observation $x_0$ of interest, rather than being restricted to samples from the joint distribution. However, reference methods are typically expensive and not infallible (e.g., they may not have converged).

Comparing two high-dimensional distributions based on a finite set of samples is difficult, and there is no single universally appropriate metric~\cite{lueckmann2021benchmarking}. The choice should be guided by the intended scientific use of the inference result. If downstream analyses rely primarily on marginal distributions, these can be compared visually or quantified through divergence measures such as the Jensen-Shannon divergence. One can also compare derived quantities such as credible/confidence intervals or maximum likelihood estimates. Alternatively, the techniques for comparing samples from one- or multi-dimensional distributions described in the following sections can be used to validate learned distributions against ground-truth samples.

The main downside of this approach is its high computational cost. As amortization is a central goal of neural simulation-based inference, validation procedures relying on non-amortized inference methods are not desirable. To address this, in the following sections, we discuss several validation methods for which samples from the joint distribution $p(x, \mu)$, such as the training and validation datasets, are sufficient, making the validation procedure amortized.

\subsection{Comparing high-dimensional distributions}\label{subsec:compare-high-dim}

Ideally, inference results should be validated by directly comparing multidimensional distributions to prevent failure conditions from being averaged out during marginalization or from remaining undetected if they are hidden in high-dimensional correlations. In many applications of simulation-based inference, the target distribution is only available through samples. For example, these could be reference posterior samples obtained using another inference method, or samples from the joint distribution generated by sampling from the likelihood. The optimal test statistic to discriminate between two distributions is given by the likelihood ratio, and can be estimated using the likelihood ratio trick introduced in Sec.~\ref{subsec:lrtrick}. In the context of validation of generative models, this is referred to as the classifier two-sample test~\cite{friedman2004multivariate,lopez2017revisiting} (C2ST). It relies on the observation that due to their simpler architecture and training task, classifiers are often able to pick up on features that were missed by generative networks, and to achieve a higher accuracy overall. Moreover, there is more flexibility in defining the classifier inputs compared to a generative model because there is no invertibility constraint. This can be exploited by passing observables that are known to be hard to learn as additional inputs to the classifier explicitly.  In its simplest form, the test compares two unconditional distributions, such as posterior samples for a fixed data point $x_0$ generated using simulation-based inference and a different reference method. In analogy to Eq.\eqref{eq:classifier_loss}, the function approximated by the classifier and the corresponding ratio estimate then read
\begin{equation}
    f_\psi(\mu) \approx \frac{p(\mu|x_0)}{q_\theta(\mu|x_0) + p(\mu|x_0)} \qquad\mand\qquad r_\psi(\mu) \approx \frac{p(\mu|x_0)}{q_\theta(\mu|x_0)} \;.
\end{equation}
The classifier output can then be evaluated using standard performance metrics for classifiers, such as the accuracy, i.e., the fraction of correctly classified samples among those classified as truth-like, or the ROC curve and the resulting area under the ROC curve (AUC). For the accuracy, there is a known analytic expression for the power of this test statistic to reject the null hypothesis of equal learned and truth distributions at a given significance~\cite{lopez2017revisiting}. For a more fine-grained result, the distribution of the ratios learned by the classifier, $r_\psi(\mu)$, can also be studied directly, and outliers can be linked to specific failure modes~\cite{Das:2023ktd}.

\begin{figure}
    \centering
    \includegraphics[width=\linewidth]{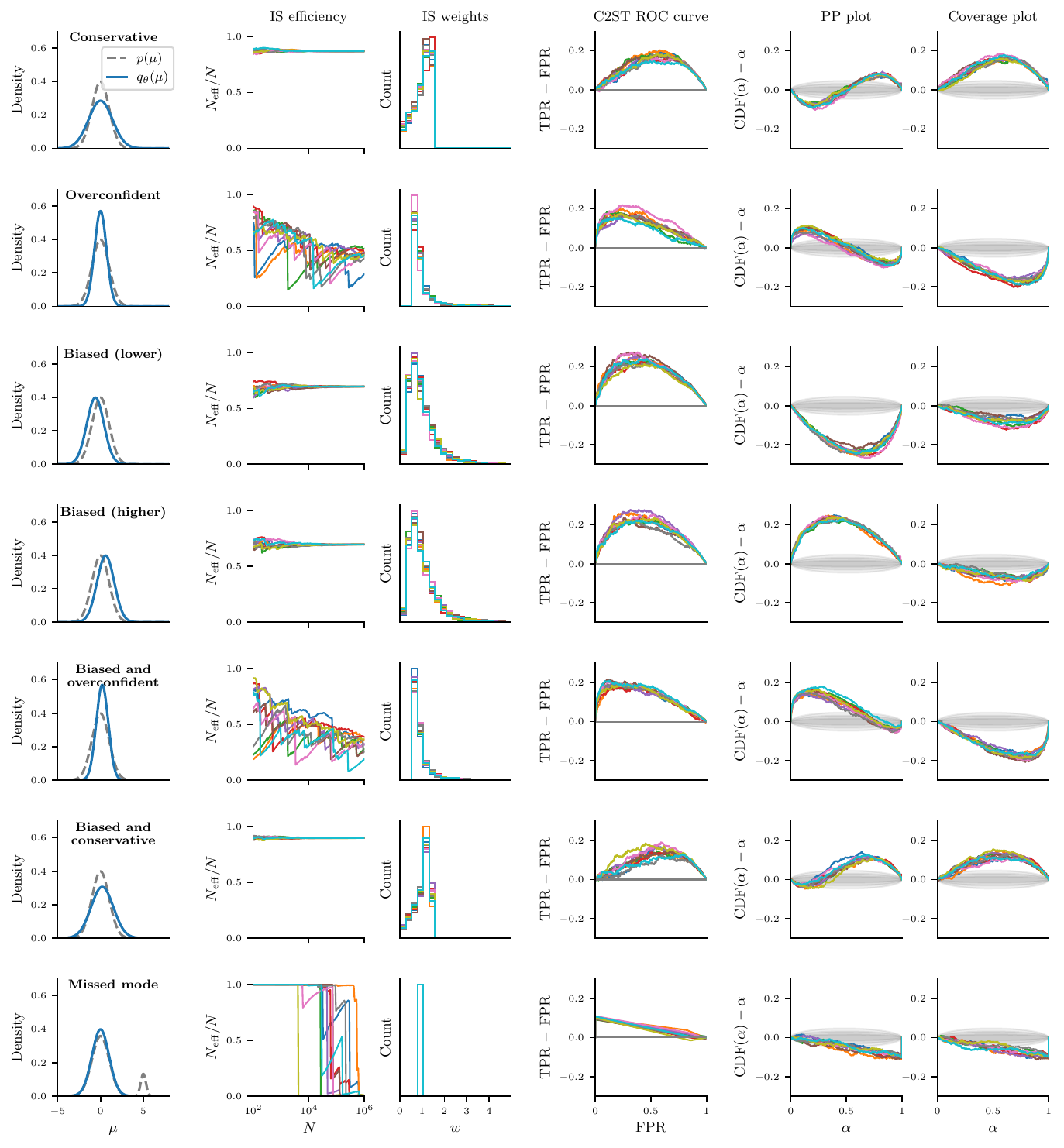}
    \caption{
    Effects of biases and under/overdispersion of a learned $q_\theta(\mu)$ (blue) compared to the target distribution $p(\mu)$ (gray) on the validation plots described in Secs.~\ref{subsec:compare-high-dim} and~\ref{subsec:calibration}. 
    We display the importance sampling efficiency as a function of the sample size $N$ (column 2), the distribution of importance weights (column 3), the related ROC curve for a C2ST test (column 4), and calibration tests in the form of a PP (column 5) or coverage (column 6) plot.
    We repeat each test 10 times (coloured lines) to illustrate statistical variations. For the tests in columns 3-6, we use a fixed number of $N=1000$ samples.
    The gray bands in the calibration plots indicate the $1/2/3\:\sigma$ levels.
    }
    \label{fig:validation}
\end{figure}

The unconditional C2ST approach is limited because it must be applied to individual data points. In local C2ST, it is extended to work with fully conditional distributions and to only require samples from the joint distribution rather than reference posterior samples from another inference method~\cite{linhart2023c2st}. Samples from the joint truth distribution can be drawn by sampling $\mu \sim p(\mu)$ from the prior and then running the simulator to sample $x$ from the likelihood, $x \sim p(x,\mu)$. Samples from the approximated joint distribution are available by using the previously sampled $x$ and drawing $\mu$ from the learned posterior, $\mu \sim q_\theta(\mu|x)$. A classifier with inputs $\mu$ and $x$ then approximates the function
\begin{equation}
    f_\psi(\mu, x) \approx \frac{p(\mu,x)}{q_\theta(\mu|x)\, p(x) + p(\mu,x)} \qquad\mand\qquad r_\psi(\mu,x) \approx \frac{p(\mu,x)}{q_\theta(\mu|x)\, p(x)} = \frac{p(\mu|x)}{q_\theta(\mu|x)}\;.
\end{equation}
Therefore, the learned classifier can be used to compare the learned posterior with the true posterior across the full data space using the same tests as for the unconditional variant. Note that the learned function $r_\psi(\mu,x)$ approximates the posterior importance sampling weights from Eq.\eqref{eq:is_weights} up to the normalization given by the evidence $p(x)$,
\begin{equation}
    r_\psi(\mu_i,x) \approx \frac{p(x|\mu_i)\, p(\mu_i)}{q_\theta(\mu_i|x)\, p(x)} = \frac{w_i}{p(x)} \;.
\end{equation}
Indeed, importance sampling can be used to correct and verify inference results when the likelihood is tractable, or its results can be approximated using classifier weights even when the likelihood is only accessible through samples. If IS achieves a large effective sample size $N_\mathrm{eff}$, this is a strong indication that the importance-sampled result is indeed reliable. Given sufficiently large $N_\mathrm{eff}$, the only remaining failure mode of importance sampling is for the network to fail to cover part of the posterior, for example, by missing a distributional mode entirely (Fig.~\ref{fig:validation} bottom row). However, such failure is inconsistent with the mass-covering training objective (Sec.~\ref{subsec:KLD-uncond}-\ref{subsec:KLD-cond}). While it can happen in practice, for example, when confronted with out-of-distribution data, the estimate should then be broadly degraded, not just in one localized region. It is therefore unlikely that the network would still provide a sufficiently accurate and heavy-tailed posterior approximation in the covered region to achieve large $N_\mathrm{eff}$. Therefore, when importance sampling yields sufficiently high $N_\mathrm{eff}$, IS provides one of the strongest available validation diagnostics in probabilistic inference. While the discussion in this section focused on neural posterior estimation, the same validation techniques can also be applied to neural likelihood estimates using a generative model, in both Bayesian and frequentist setups.

We illustrate the sampling efficiency $N_\mathrm{eff} / N$ and the distribution of self-normalized importance weights as well as the corresponding classifier ROC curves for different failure modes in Fig.~\ref{fig:validation}. A conservative estimate $q_\theta$ (rows 1 and 6 in Fig.~\ref{fig:validation}) yields stable sample efficiencies, with the weight distribution peaking above one with a long tail towards low weights. An overconfident (rows 2, 5, 7) estimate results in unstable efficiency, with a weight distribution peaking below one and a tail towards large weights. These high weights are associated with the light tails of the distribution, where $q_\theta$ drastically underestimates $p$. 
Biases (rows 3-6) yield degraded sample efficiency and a widened weight distribution, both towards large and small weights. 
In all cases, the ROC curve shows a significant deviation from the ideal linear ROC curve, but does not allow for the identification of specific failure modes.

\subsection{Calibration tests}\label{subsec:calibration}

Simulation-based calibration~\cite{talts2018validating} is a popular approach for validating whether a modeled distribution $q_\theta$ agrees with a target distribution $p$. Because it is limited to one-dimensional distributions, a set of observables for calibration must be selected first. In statistics, such functions mapping the full parameter space to a one-dimensional number are referred to as pre-rank functions~\cite{gneiting2008assessing},
\begin{equation}
    f\::\:\mu\mapsto f(\mu)\in \mathbb{R} \; .
\end{equation}
They can include marginal distributions from the input space the network was trained on, as well as observables that encode hard-to-learn correlations. Defining a suitable set of observables usually requires domain knowledge. The next step is to define a transformation that converts $q_\theta$ into a univariate uniform distribution. This transformation $F_{f,q_\theta}$ is called probability integral transform (PIT)~\cite{rosenblatt1952remarks}, and is given by the cumulative distribution function of $f(\mu)$ computed from samples $\mu \sim q_\theta(\mu)$,
\begin{equation}
    F_{f,q_\theta}(f(\mu)) = P_{\mu' \sim q_\theta(\mu')}\big\{ f(\mu') \leq f(\mu) \big\} \;.
\end{equation}
Consequently, the transformed samples are uniformly distributed,
\begin{equation}
    p(F) = \text{const.} \qquad\mfor\qquad F = F_{f,q_\theta}(f(\mu))
    \qquad\mwith\qquad \mu \sim q_\theta(\mu) \;.
\end{equation}
The next step is to apply the same transformation to samples from the target distribution, $\mu \sim p(\mu)$. If $q_\theta=p$, these transformed samples should also follow a uniform distribution. A finite set of samples,
\begin{equation}
    \{F_1, \ldots, F_n\} \equiv \{F_{f,q_\theta}(f(\mu_1)),\ldots,F_{f,q_\theta}(f(\mu_n))\}
    \qquad\mwith\qquad \mu_i \sim p(\mu) \;,
\end{equation}
can be compared to a uniform distribution graphically~\cite{sailynoja2022graphical} by plotting the function,
\begin{equation}
    U(F) = \frac{1}{n}\sum_{i=1}^n \mathbb{I}(F_i < F) \qquad\mfor\qquad F \in [0,1] \;,
\end{equation}
corresponding to the CDF for $F$ estimated from a finite set of samples. In practice, this function can be easily plotted by sorting the $F_i$ in ascending order and then drawing a step function through the points $(F_i, i/n)$. Ideally, it should match the identity function. Alternatively, the discrepancy between the distributions can be quantified using test statistics such as the Kolmogorov-Smirnov score,
\begin{equation}
    D_\text{KS} = \sup_{F \in [0,1]}\left|U(F) - F\right|\;.
\end{equation}
Visually, it corresponds to the largest vertical difference between the approximated CDF $U(F)$ and the identity function in the plot above. The statistical significance of such calibration tests increases with $n$. With sufficiently large $n$, even small deviations between $q_\theta$ and $p$ can be identified with high statistical significance.

For simulation-based inference, we can usually not directly sample the target distribution $p(\mu|x)$. However, sampling from the joint distribution $(\mu_i,x_i)\sim p(\mu)p(x|\mu)$ produces one single posterior sample $\mu_i$ for the corresponding observation $x_i$ (Sec.~\ref{subsec:KLD-cond}). While a calibration test with a single sample is not useful on its own, one can compensate by applying it to a set of $n$ different observations. This proceeds analogously to the tests for unconditional distributions as described above, but for each parameter $\mu_i\sim p(\mu)$, one additionally samples the corresponding observation from the likelihood $x_i\sim p(x|\mu_i)$, and evaluates the PIT with respect to the corresponding conditional estimate $q_\theta(\mu|x_i)$. In the following, we discuss PP plots, coverage plots, and classifier calibration plots as three important examples of such calibration tests in simulation-based inference.

\subsubsection{Probability-probability (PP) plots}

A common calibration test is the so-called probability-probability (PP) plot, for which the pre-rank function projects onto a 1D marginal distribution. For all $n$ samples from $p$, the CDF under the marginalized $q_\theta$ has to be computed. In an amortized inference setup with a generative network, this can be done cheaply by drawing $m$ samples from  $q_\theta$ and estimating the CDF similarly from the samples using 
\begin{equation}
    F_{f,q_\theta}(f(\mu_i)) \approx \frac{1}{m}\sum_{j=1}^m \mathbb{I}( f(\mu'_j) \leq f(\mu_i) )
    \qquad\mwith\qquad \mu'_j \sim q_\theta(\mu'_j) \;.
\end{equation}
The calibration curve is then plotted as described above. Intuitively, this tests (for the corresponding marginal) whether a sample from $p$ has a probability of $\alpha$ of falling below the $\alpha$-th quantile of $q_\theta$ for any $\alpha\in[0,1]$. PP plots can be used for posterior and likelihood estimation, and are therefore applicable in both the Bayesian and frequentist frameworks when a generative model is used.

\begin{figure}
    \centering
    \includegraphics[width=\linewidth]{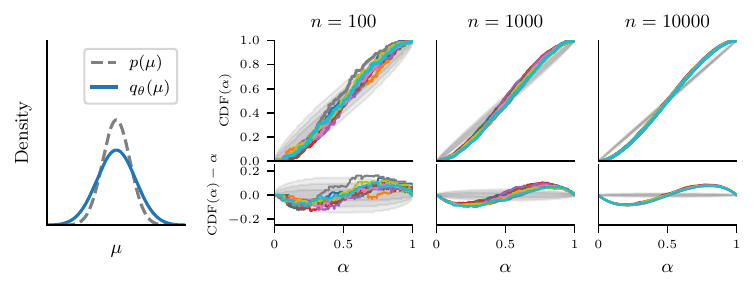}
    \caption{
        PP plots for a conservative estimate $q_\theta$ of a Gaussian target $p$. The PP test has intrinsic statistical uncertainty inherited from sampling $\mu\sim p(\mu)$ (10 independent tests shown in coloured lines), which decreases with increasing number of samples $n$. The gray bands in the calibration plots indicate the $1/2/3\:\sigma$ levels.
    }
    \label{fig:PP-plot}
\end{figure}

We illustrate how typical failure modes show up in the PP plot in Fig.~\ref{fig:validation}. Overconfident distributions lead to an S-shaped calibration curve that is initially too high, then crosses the diagonal in the middle and is eventually too low. This is because samples from $p$ frequently fall in the tails of the too-narrow $q_\theta$. The transformed values thus accumulate near 0 and 1. For a conservative $q_\theta$ on the other hand, the shape of the curve flips: it is initially too low and eventually too high, as the transformed values cluster around 0.5. In the case of a bias, the cumulative distribution for $q_\theta$ is above or below the one for $p$ everywhere, leading to a curved shape. For a perfectly calibrated distribution, we expect that at a point $\alpha$ in the calibration curve for a given sample, the transformed quantity should be below $\alpha$ with a probability $\alpha$. For $n$ samples, the value of the calibration curve is therefore described by a binomial distribution,
\begin{equation}
    P\left(\frac{i}{n}\right) = \binom{n}{i} \alpha^i (1-\alpha)^{n-i} \;.
\end{equation}
This can be used to decide whether a given deviation from the diagonal is significant or due to fluctuations arising from the limited sample size $n$. This statistical uncertainty can be visualized through bands for the different $\sigma$ intervals (Fig.~\ref{fig:PP-plot}).

\subsubsection{Coverage plots}

Similarly, one can test whether a sample falls in the $\alpha$ credible region of $q_\theta$ with probability $\alpha$, yielding a so-called coverage plot. Such coverage plots can be made for marginals and for the full multidimensional distribution. Unlike PP plots, coverage plots are not limited to one-dimensional observables and can equivalently be defined for multi-dimensional $\alpha$-credible sets. Estimating coverage in one dimension, however, is typically computationally cheaper. By treating $x$ as the random quantity for a fixed $\mu$, and replacing credible sets with confidence sets, coverage plots can also be used in a frequentist inference setup in combination with likelihood or ratio estimation.

We again illustrate how different failure modes affect the coverage plots in Fig.~\ref{fig:validation}. Compared to PP plots, coverage plots are much more sensitive to overconfident or conservative distributions, leading to a calibration curve that lies entirely above or below the diagonal. The effect of bias on the coverage plots is more subtle, with smaller deviations from the diagonal. Therefore, PP plots and coverage plots can be considered complementary tools for identifying various failure modes.

\subsubsection{Classifier calibration plots}

So far, our discussion has focused on comparing samples from two probability distributions. The same framework can be applied to neural ratio estimation. Consider the example of learning the likelihood-to-evidence ratio as introduced in Sec.~\ref{subsec:nre}. Following Eq.\eqref{eq:classifier_loss}, a classifier trained to discriminate between samples from the joint distribution $(x,\mu) \sim p(x,\mu)$ with label 1 and shuffled samples $(x,\mu) \sim p(x)\,p(\mu)$ with label 0 approximates the function
\begin{equation}
    f_\psi(x|\mu) \approx \frac{p(x,\mu)}{p(x,\mu) + p(x)\,p(\mu)} \;. 
\end{equation}
For a perfect classifier, the function $f_\psi(x|\mu) \in [0,1]$ quantifies the probability that a sample $(x,\mu)$ from the mixed distribution $p(x,\mu) + p(x)\,p(\mu)$ was drawn from the joint distribution $p(x,\mu)$. The corresponding cumulative distribution can also be estimated by averaging over the labels in the validation dataset. Therefore, the classifier output can be used directly to create a calibration plot by extracting the corresponding PIT, for example, using quantiles of the validation data.

Beyond their use as a diagnostic tool, calibration curves for classifiers can also be used to define a monotonic transformation that improves the calibration of the classifier by applying it to the classifier output as a postprocessing step~\cite{Cranmer:2015bka}. Common methods for extracting such a transformation from samples include isotonic regression and quantile regression. This calibration step is possible even when the likelihood ratio learned by the classifier is a poor approximation of the true likelihood ratio. In such cases, it results in a loss of statistical power while improving the validity of confidence sets.

\subsubsection{Limitations}

For univariate unconditional distributions ($\mu\in\mathbb{R}$), calibration tests are a sufficient condition for $q_\theta(\mu)=p(\mu)$ in the limit $n\to\infty$. For multivariate unconditional distributions ($\mu\in\mathbb{R}^m$, $m>1$), they are necessary but not sufficient, as the pre-rank functions are lossy. 
For conditional distributions, calibration tests are necessary but not sufficient to check $q_\theta(\mu|x) = p(\mu|x)$ even in the univariate case. This is because calibration tests marginalize over observations $x_i$, such that biases for different $x_i$ could cancel each other. For example, a model $q_\theta(\mu|x)=p(\mu)$ simply estimating the prior would perfectly pass any joint calibration test, despite being a poor approximation to $p(\mu|x)$. To address these problems, methods have been proposed to define sufficient coverage tests~\cite{lemos2023sampling,lemos2024pqmass}.

Calibration tests are universally applicable to validate simulation-based inference, as they only require likelihood simulations. They thus serve as relatively cheap sanity checks. On the other hand, they are subject to several limitations. 
First, they do not provide sufficient conditions for accurate inference.
Second, they are restricted to simulated observations, generated with parameters sampled from the training prior. This impedes calibration tests with real observed data or targeted tests in specific parameter regions of interest.
Third, marginalization over large sets of observations makes it difficult to analyze failure cases, as a biased calibration plot does not reveal detailed information about individual observations.

\subsection{Posterior predictive checks}\label{subsec:posterior-predictive-checks}
The diagnostics discussed above operate in parameter space and, in the case of calibration tests, are restricted to simulated observations. Posterior predictive checks instead operate in data space and can be applied directly to a real observation $x_0$. One first draws samples from the estimated posterior $\mu_i\sim q_\theta(\mu|x_0)$, and then runs the simulator for each of them, $x_i\sim p(x|\mu_i)$. If inference is accurate and the simulator is consistent with the data, $x_0$ should be typical of the posterior predictive samples $\{x_i\}$, which is usually assessed in terms of domain-informed test statistics~\cite{gelman1996posterior,gelman2020bayesian}. A failed check could indicate either inaccurate inference or a simulator that is unable to generate the observed data. While posterior predictive checks do not distinguish between these two failure modes, they are one of the few diagnostics sensitive to simulator misspecification. Furthermore, they are always applicable in simulation-based inference, as they only require simulator samples. On the other hand, posterior predictive checks rely on the availability of meaningful test statistics and passing them does not provide a sufficient validation criterion.

\clearpage
\section{Outlook}\label{sec:outlook}
Simulation-based inference with machine learning has seen widespread adoption across scientific disciplines in recent years. The core idea is to train an inference network on samples from the joint distribution $p(\mu,x)$ over parameters and observations (Sec.~\ref{subsec:ml-gen-models}--\ref{subsec:ml-classifiers}). This setup is compatible with both Bayesian and frequentist perspectives (Sec.~\ref{subsec:stat-bayes}--\ref{subsec:stat-freq}), and extends naturally to distribution-level tasks such as empirical Bayes and unfolding (Sec.~\ref{subsec:ml-emp-bayes}). It is particularly useful when the likelihood is intractable, while a tractable likelihood, if available, can still be exploited during training or inference (Sec.~\ref{subsec:ml-likelihood}). Once trained, a network often enables fast, amortized inference on new observations at minimal cost.

The most important open challenge is the lack of universal validation techniques. Almost all tractable diagnostics (Sec.~\ref{sec:verification}) are necessary but not sufficient conditions for accurate inference: a diagnostic can pass even if the inference result is incorrect in ways that matter for downstream science. This limitation is particularly critical in applications where no alternative method is available as a reference.
Another major challenge is simulator misspecification~\cite{cannon2022investigating,schmitt2023detecting,wehenkel2024addressing} (see also the dedicated VERaiPHY article on this topic~\cite{veraiphy_misspecification}). Even a well-trained network may produce inaccurate results for observations that are incompatible with the simulator, since neural networks typically generalize poorly to out-of-distribution data. In such cases, likelihood-based methods can be more robust (provided the likelihood remains well-defined, i.e., the observation is improbable rather than impossible under the model). 

Both these limitations are major impediments to adoption in scientific practice. Indeed, most work on simulation-based inference to date has focused on developing methods and adapting them to domain-specific problems, while analyses that yield new scientific results remain comparatively rare (though this is beginning to change, with frameworks deployed in large experiments~\cite{Dax:2021tsq}, and first SBI-driven results~\cite{ATLAS:2025clx,Gupte:2024jfe} emerging). This partly reflects the relative youth of the field, but also the absence of established validation standards and statistically robust methods to deal with misspecified simulators.

Further open challenges include the treatment of large numbers of nuisance parameters and reliable estimation of the Bayesian evidence for model selection. Simulation-based methods for both remain less mature than those for parameter estimation.

Beyond these methodological questions, simulation-based inference also faces the practical challenge of being inherently interdisciplinary. Relevant developments are thus scattered across application domains, which can make it difficult for practitioners to identify the most suitable tools. Dedicated software packages~\cite{BoeltsDeistler_sbi_2025,rozet2021lampe,Miller:2022shs,bayesflow_2023_software} aim to lower this barrier by providing accessible implementations. At the same time, scientific simulators are increasingly being implemented in machine learning frameworks such as PyTorch~\cite{paszke2019pytorch} and JAX~\cite{jax2018github}, providing native GPU support and automatic differentiation. Simulation-based inference stands to benefit directly from this trend, for instance by leveraging simulator gradients during training.

As experiments across many domains produce ever-larger datasets while simulators remain computationally expensive, efficient and amortized simulation-based inference will become an increasingly important tool for scientific discovery.

\clearpage

\section*{Acknowledgements}

We thank Oz Amram, Manuel Haußmann, and Sofia Palacios Schweitzer for their valuable feedback. Furthermore, we thank Tilman Plehn, Lydia Brenner, and Louis Lyons for starting the VERaiPHY initiative, and Ramon Winterhalder and Gaia Grosso for coordinating the effort.
TH is supported by the PDR-Weave grant FNRS-DFG numéro T019324F (40020485), and by FRS-FNRS (Belgian National Scientific Research Fund) IISN projects 4.4503.16 (MaxLHC).

\bibliography{refs}
\end{document}